%% file: main.tex
\documentclass[10pt,twocolumn,letterpaper]{article}

\usepackage[pagebackref=true,breaklinks=true,colorlinks,bookmarks=false]{hyperref}

\usepackage{cvpr}
\usepackage{times}
\usepackage{epsfig}
\usepackage{graphicx}
\usepackage{amsmath}
\usepackage{amssymb}
\usepackage{array}
\usepackage{gensymb}
\usepackage{nicefrac}
\usepackage{tabularx}
\usepackage{booktabs}

\cvprfinalcopy 

\def\cvprPaperID{2618} 

\begin{document}

\title{A Perceptual Measure for Deep Single Image Camera Calibration\vspace{-1em}}


\author{Yannick Hold-Geoffroy\textsuperscript{1*},\: Kalyan Sunkavalli\textsuperscript{$\dagger$},\: Jonathan Eisenmann\textsuperscript{$\dagger$},\: Matt Fisher\textsuperscript{$\dagger$},\\ Emiliano Gambaretto\textsuperscript{$\dagger$},\: Sunil Hadap\textsuperscript{$\dagger$},\: Jean-Fran\c{c}ois Lalonde\textsuperscript{*}\\
Universit\'e Laval\textsuperscript{*}, Adobe\textsuperscript{$\dagger$}\\
{\tt\scriptsize yannick.hold-geoffroy.1@ulaval.ca, \{sunkaval,eisenman,matfishe,emiliano,hadap\}@adobe.com, jflalonde@gel.ulaval.ca}\vspace{-1em}}

\maketitle
\thispagestyle{empty}

\begin{abstract}
   Most current single image camera calibration methods rely on specific image features or user input, and cannot be applied to natural images captured in uncontrolled settings. We propose directly inferring camera calibration parameters from a single image using a deep convolutional neural network. This network is trained using automatically generated samples from a large-scale panorama dataset, and considerably outperforms other methods, including recent deep learning-based approaches, in terms of standard L2 error. However, we argue that in many cases it is more important to consider how humans perceive errors in camera estimation. To this end, we conduct a large-scale human perception study where we ask users to judge the realism of 3D objects composited with and without ground truth camera calibration. Based on this study, we develop a new perceptual measure for camera calibration, and demonstrate that our deep calibration network outperforms other methods on this measure. Finally, we demonstrate the use of our calibration network for a number of applications including virtual object insertion, image retrieval and compositing.

\end{abstract}

\makeatletter
\def\blfootnote{\gdef\@thefnmark{}\@footnotetext}

\renewcommand{\paragraph}{%
  \@startsection{paragraph}{4}%
  {\z@}{1.25ex \@plus 1ex \@minus .2ex}{-1em}%
  {\normalfont\normalsize\bfseries}%
}

\makeatother

\blfootnote{\scriptsize \textsuperscript{1} Research partly done when Y. Hold-Geoffroy was an intern at Adobe Research.}


\input{intro}

\input{relatedwork}
\input{representation}
\input{method}

\input{pstudy}

\input{method_evaluation_userstudy}
\input{applications}

\input{discussion}

\clearpage

\section{Acknowledgments}

Parts of this work were done while Yannick Hold-Geoffroy was an intern at Adobe Research. This work was partially supported by the REPARTI Strategic Network and the NSERC Discovery Grant RGPIN-2014-05314. We gratefully acknowledge the support of Nvidia with the donation of the GPUs used for this research.

{\small
\bibliographystyle{ieee}
\bibliography{refs}
}

\end{document}

%% file: intro.tex

\section{Introduction}


The first step for many vision and graphics tasks---ranging from 3D scene reconstruction to image metrology to photographic editing---is to geometrically calibrate the camera that captured the image~\cite{Hartley2004}. In this work, we are specifically interested in calibrating a camera from a single image of a natural scene, thereby precluding the use of multiple views of the scene or calibration targets. There is a extensive body of work even in this challenging setting; however, most approaches rely on detecting specific image features like vanishing lines~\cite{Lee2014}, coplanar circles~\cite{Chen2004} and repeated texture patterns~\cite{Schaffalitzky2000,Criminisi00,Pritts2014}, making them inapplicable to images without these features.

Inspired by the success of deep learning for related scene reconstruction tasks~\cite{choy20163d,bianco2017single}, we propose training a deep convolutional network to directly estimate camera parameters---more specifically the focal length, pitch and roll--from a single image. This is similar to recent work on CNN-based focal length~\cite{Workman2015a} and horizon estimation~\cite{Workman2016}. We significantly improve on their results by jointly estimating all the parameters and by training on sample images automatically extracted from a large-scale panorama dataset~\cite{Xiao2012}. We also analyze the features learned by the network to understand how it differs from pre-defined feature-based calibration.

While our calibration network produces state-of-the-art results, there are a number of cases where it fails. For images with no clear geometric or semantic cues, there is ambiguous evidence for what the ``right'' camera calibration parameters should be. In many situations, recovering the exact camera calibration may not even be required; for example, humans are known to tolerate strong deviations from realism in painting~\cite{Cavanagh2005} and digital composites~\cite{Farid2010}. This lead us to ask the question: \emph{how do humans perceive inaccuracies in geometric camera calibration}? To answer this, we conducted a large-scale user study to test the human perception of errors in camera calibration. We composited virtual objects into images using both ground truth calibration parameters as well as randomly perturbed parameters, and asked users to evaluate which image in each pair looked ``real'' to them. The results of this user study are described in sec.~\ref{sec:human_sensitivity_analysis}. To our knowledge, this is the first large-scale user study to systematically evaluate human perception of camera perspective. We use this data to design a new perceptual measure for calibration errors that we believe can prove to be important as we evaluate how precise our algorithms need to be for applications like augmented reality and 3D object composition.

While our calibration network was trained to minimize an entropy-based loss, we show that it also outperforms previous methods on our perceptual measure. In addition, we also demonstrate the use of our network for a wide range of applications, including 3D object insertion, calibration-based image retrieval and compositing.  

\newcommand{\exampleresultswidth}{0.195}
\begin{figure*}
\centering
\bgroup
\setlength{\tabcolsep}{1pt}
\begin{tabular}{cccc|c}
\includegraphics[width=\exampleresultswidth\linewidth]{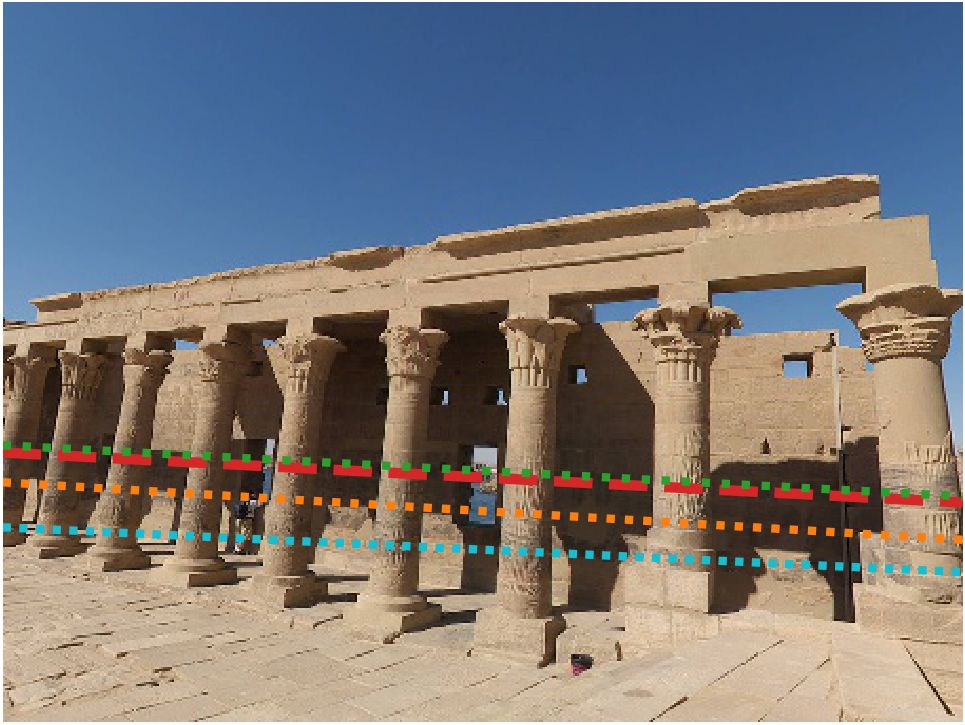} &
\includegraphics[width=\exampleresultswidth\linewidth]{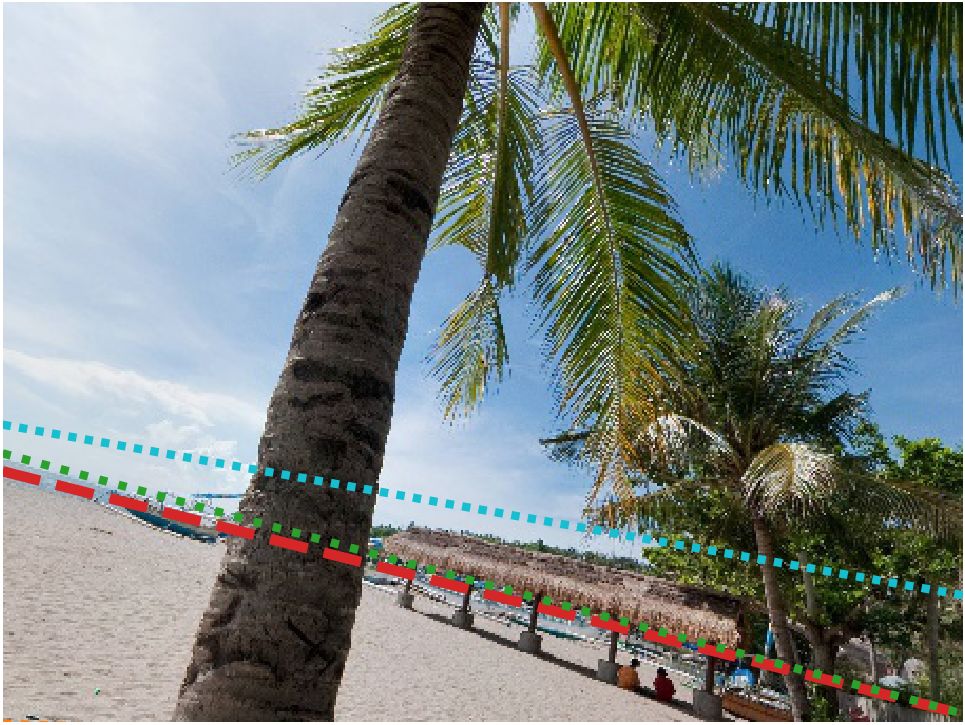} &
\includegraphics[width=\exampleresultswidth\linewidth]{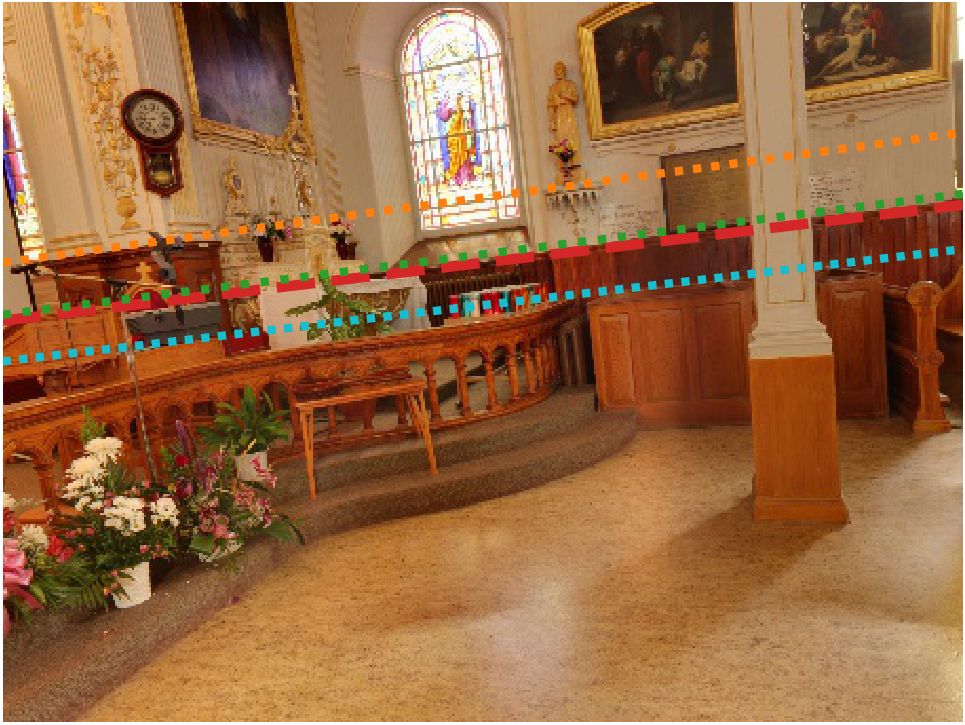} &
\includegraphics[width=\exampleresultswidth\linewidth]{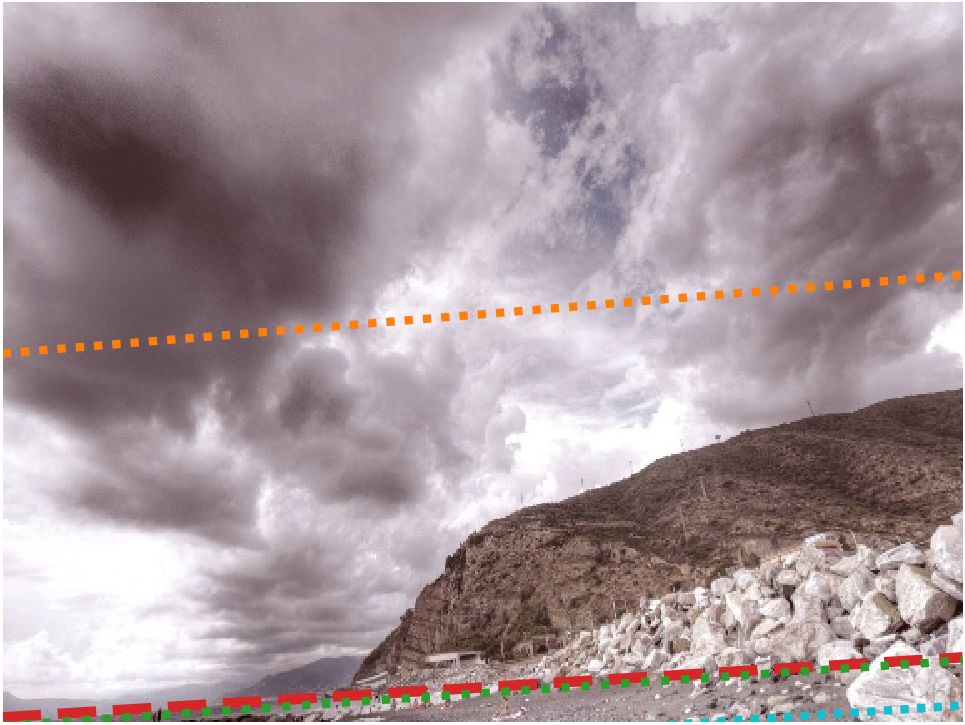} &
\includegraphics[width=\exampleresultswidth\linewidth]{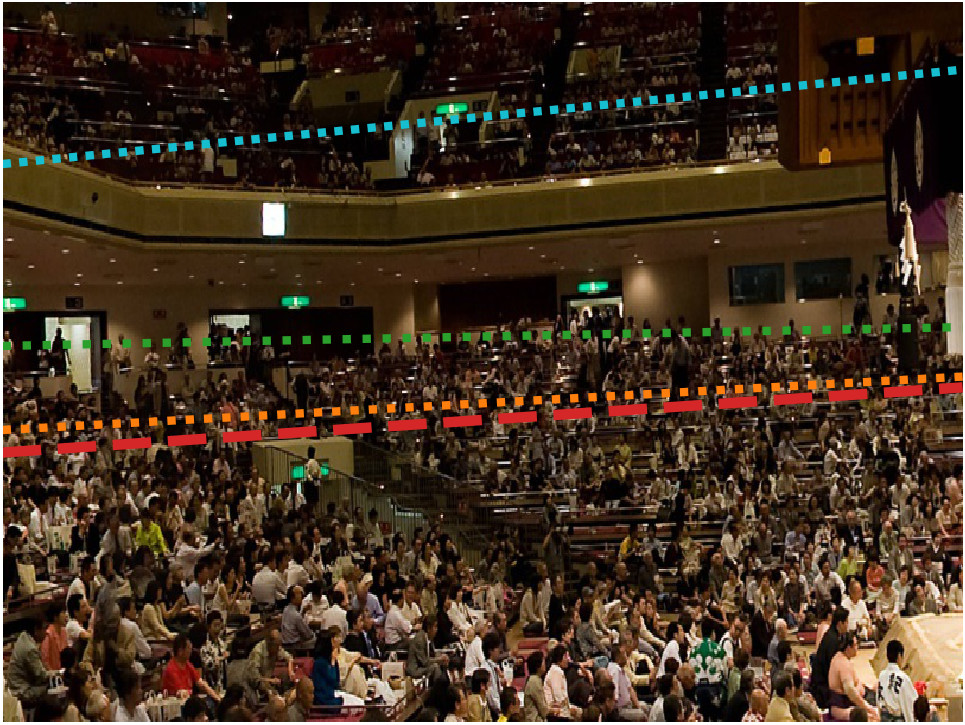} \\
\includegraphics[width=\exampleresultswidth\linewidth]{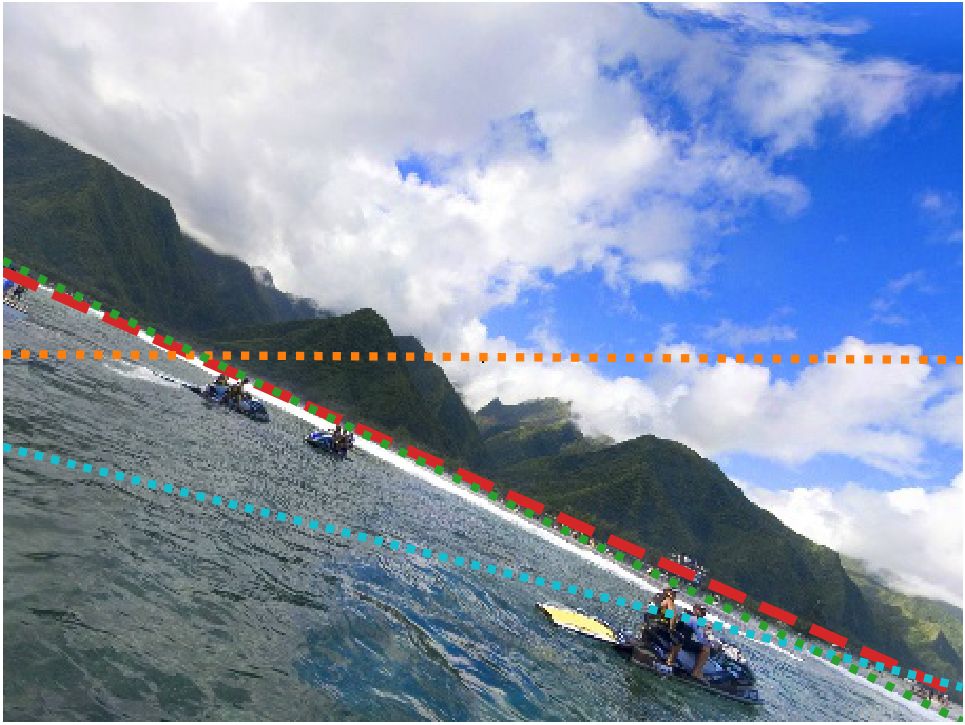} &
\includegraphics[width=\exampleresultswidth\linewidth]{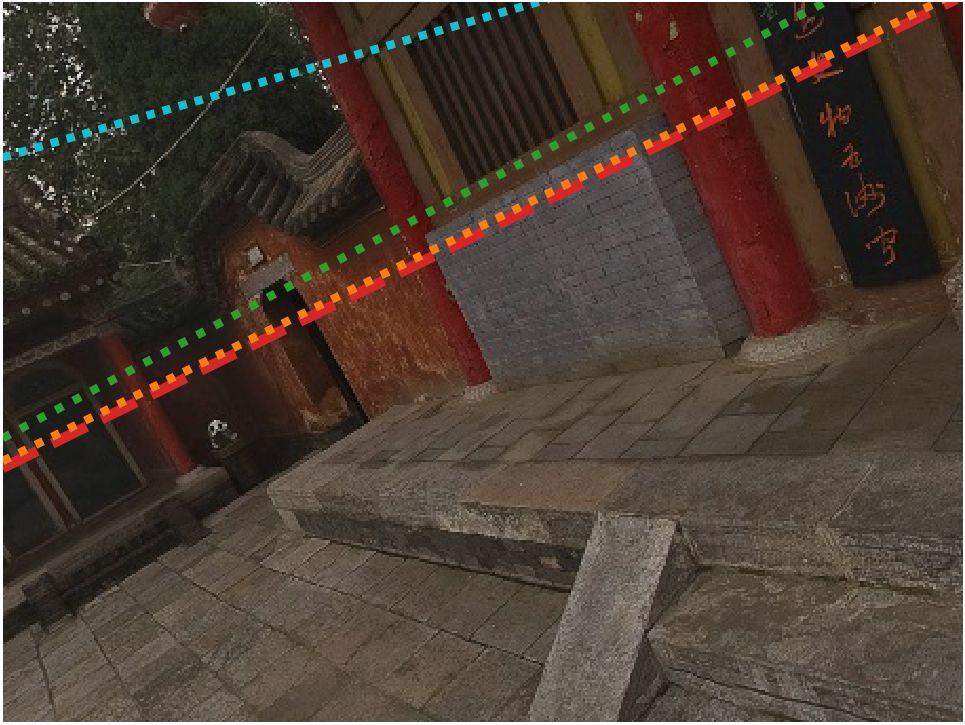} &
\includegraphics[width=\exampleresultswidth\linewidth]{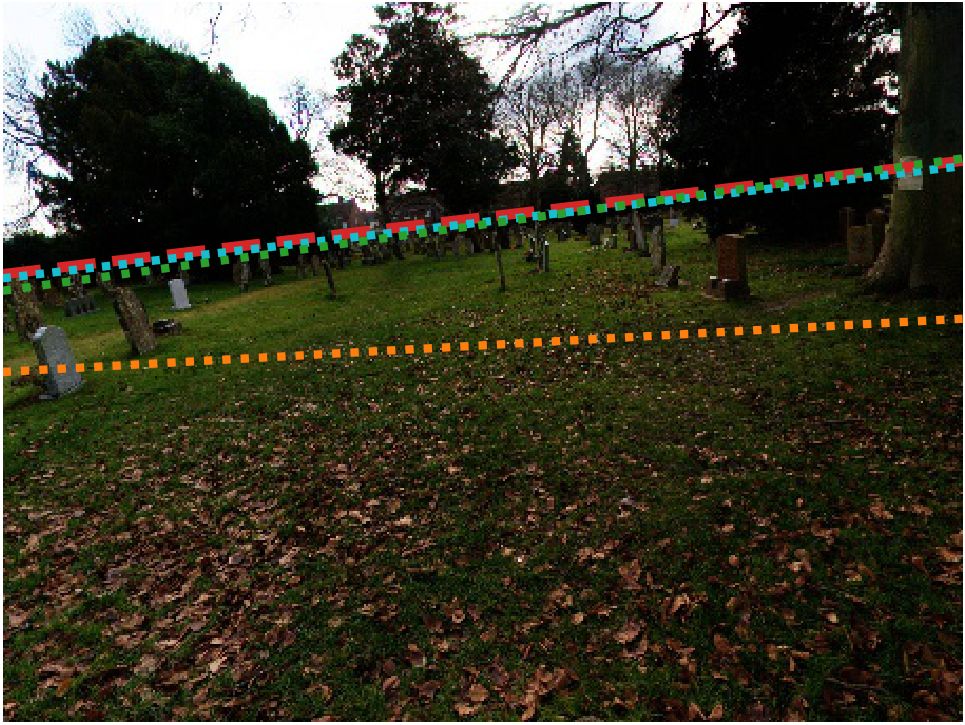} &
\includegraphics[width=\exampleresultswidth\linewidth]{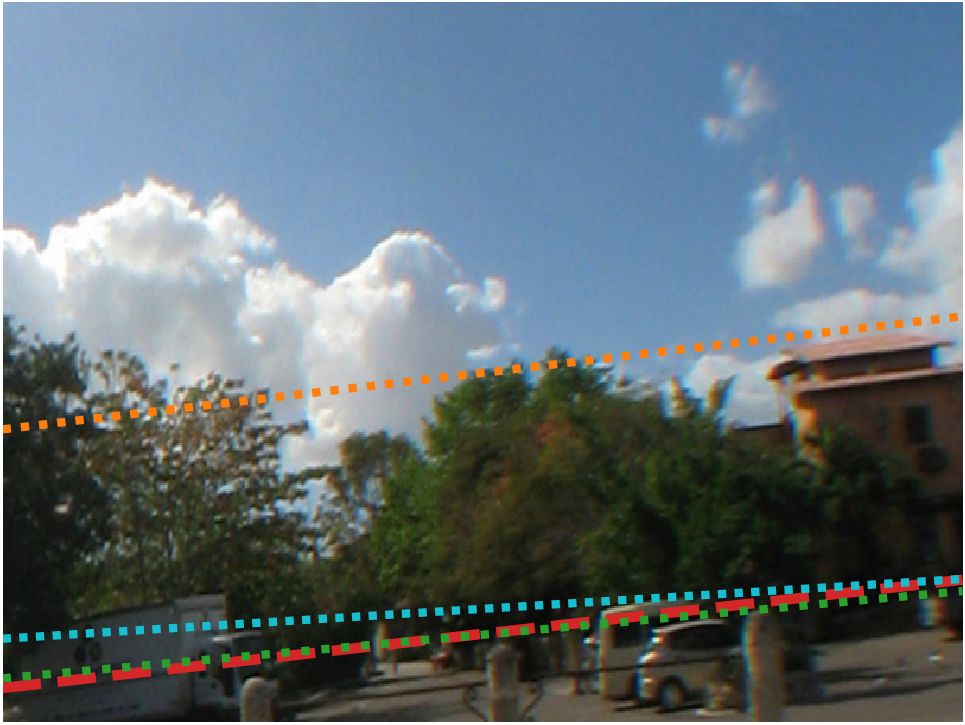} &
\includegraphics[width=\exampleresultswidth\linewidth]{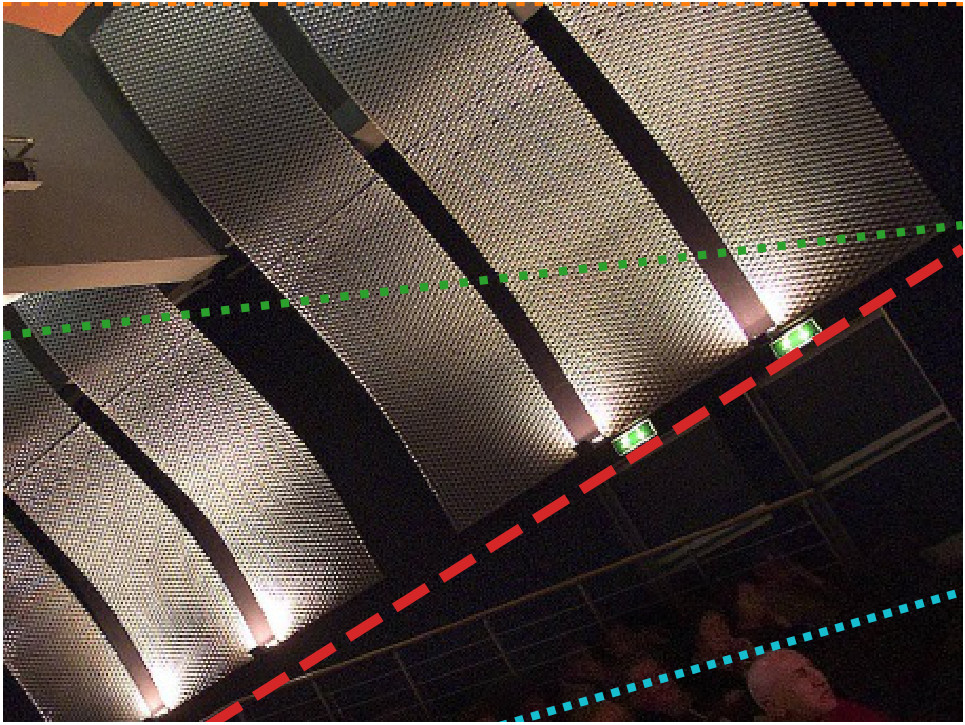} \\
\includegraphics[width=\exampleresultswidth\linewidth]{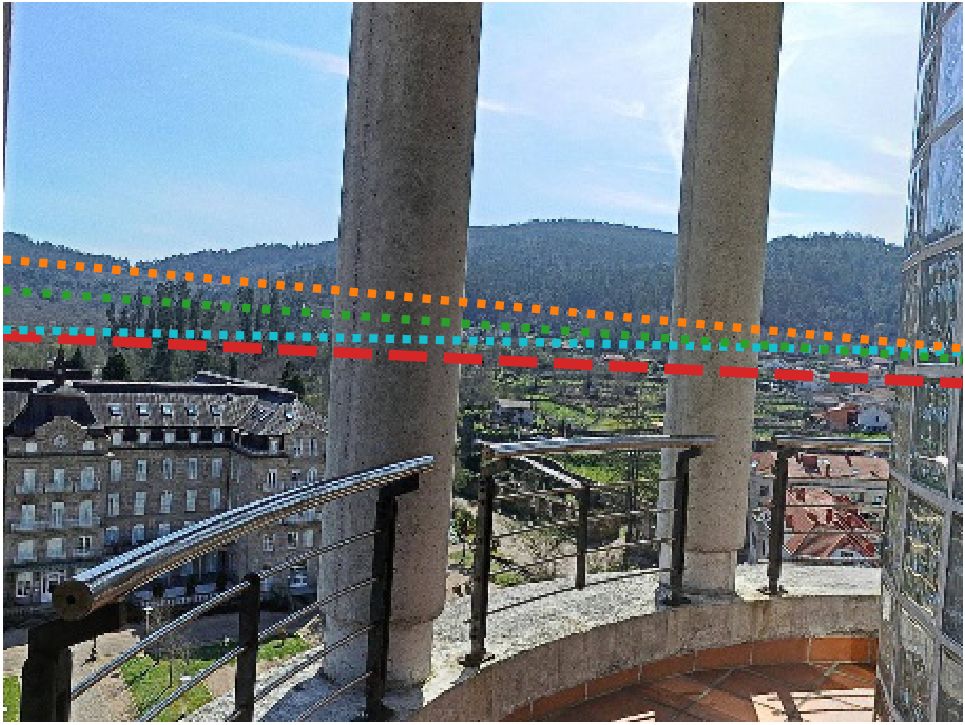} &
\includegraphics[width=\exampleresultswidth\linewidth]{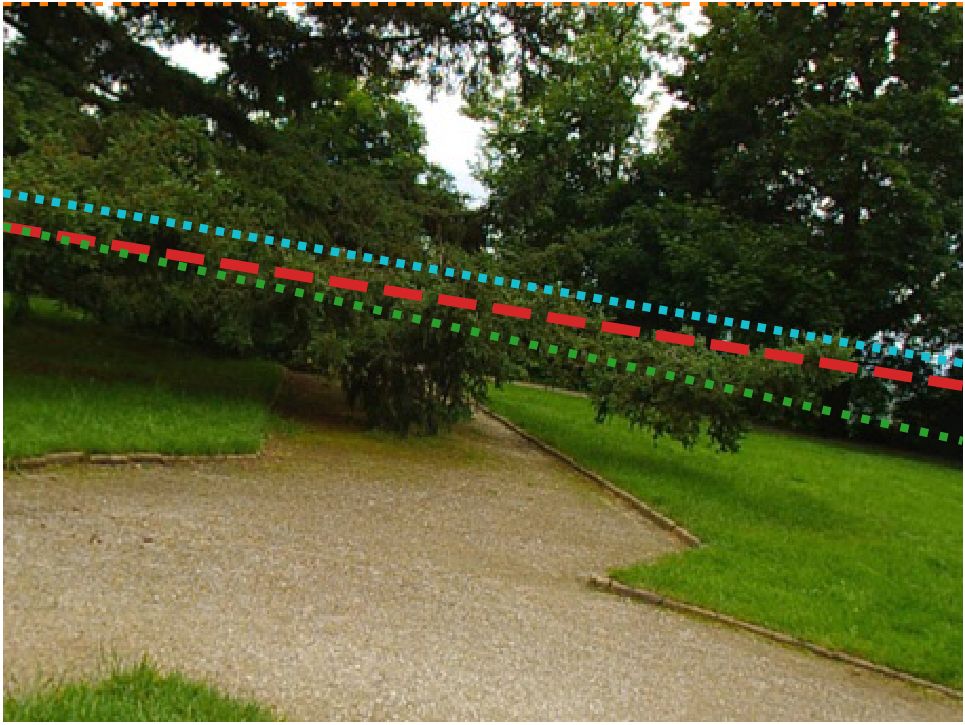} &
\includegraphics[width=\exampleresultswidth\linewidth]{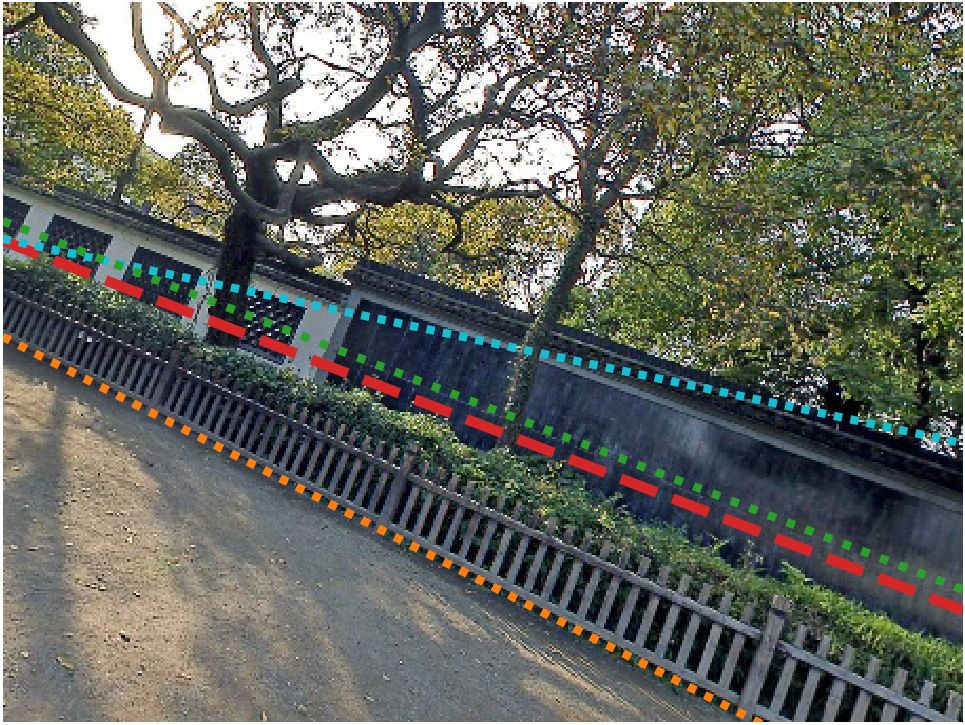} &
\includegraphics[width=\exampleresultswidth\linewidth]{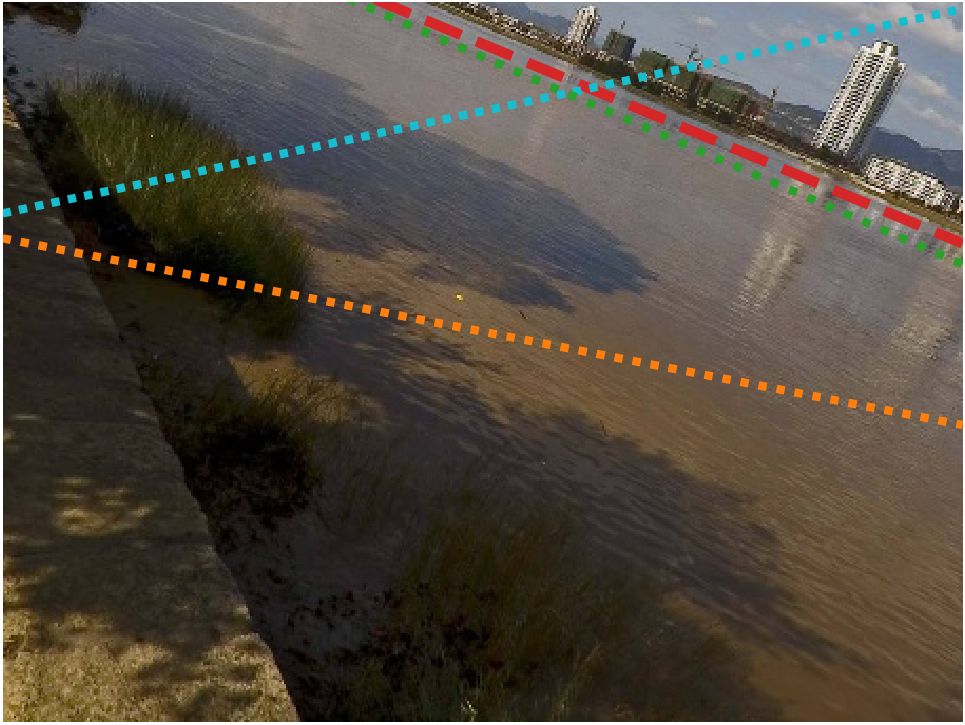} &
\includegraphics[width=\exampleresultswidth\linewidth]{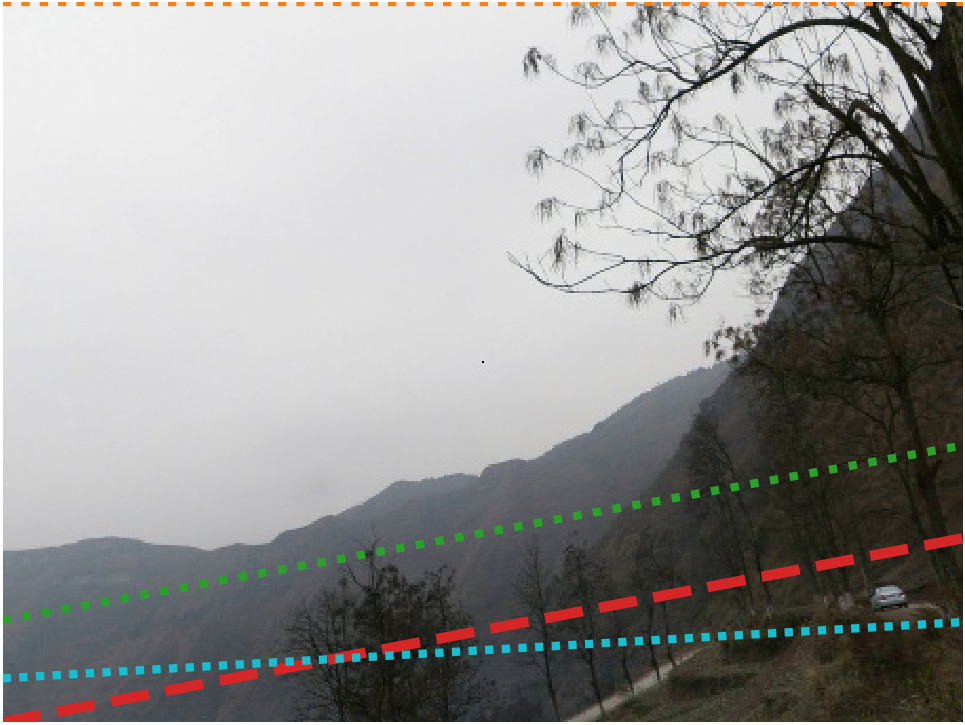} \\
\includegraphics[width=\exampleresultswidth\linewidth]{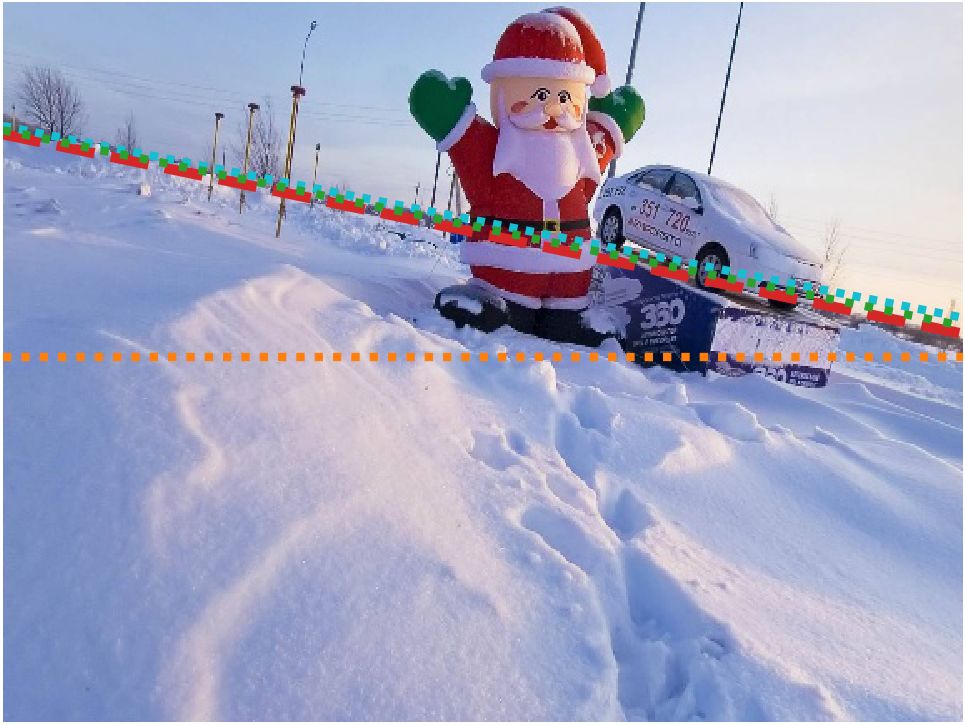} &
\includegraphics[width=\exampleresultswidth\linewidth]{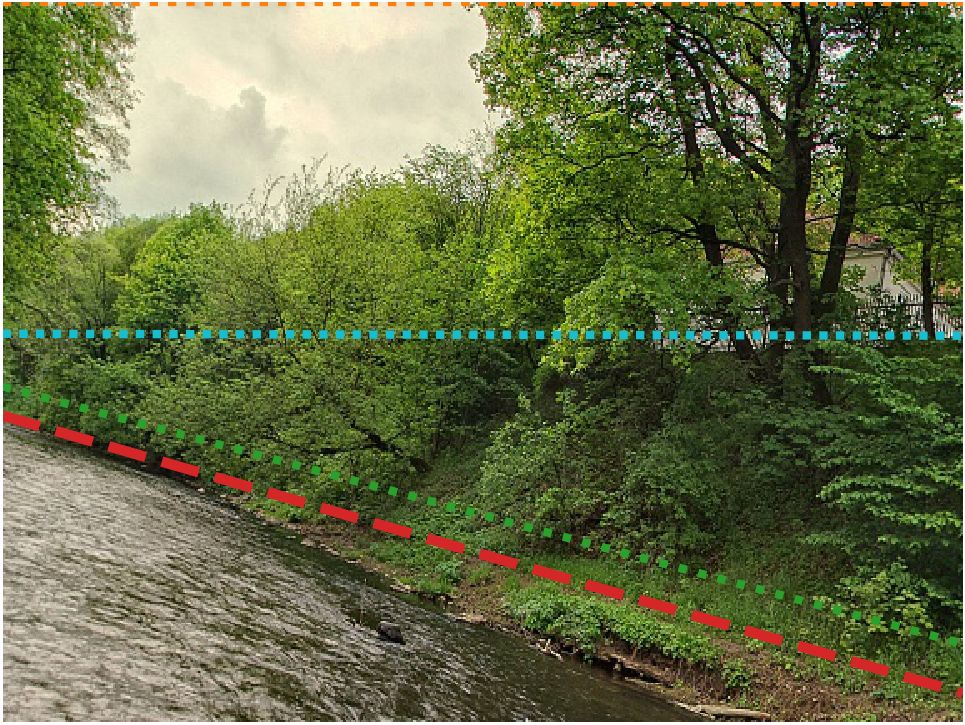} &
\includegraphics[width=\exampleresultswidth\linewidth]{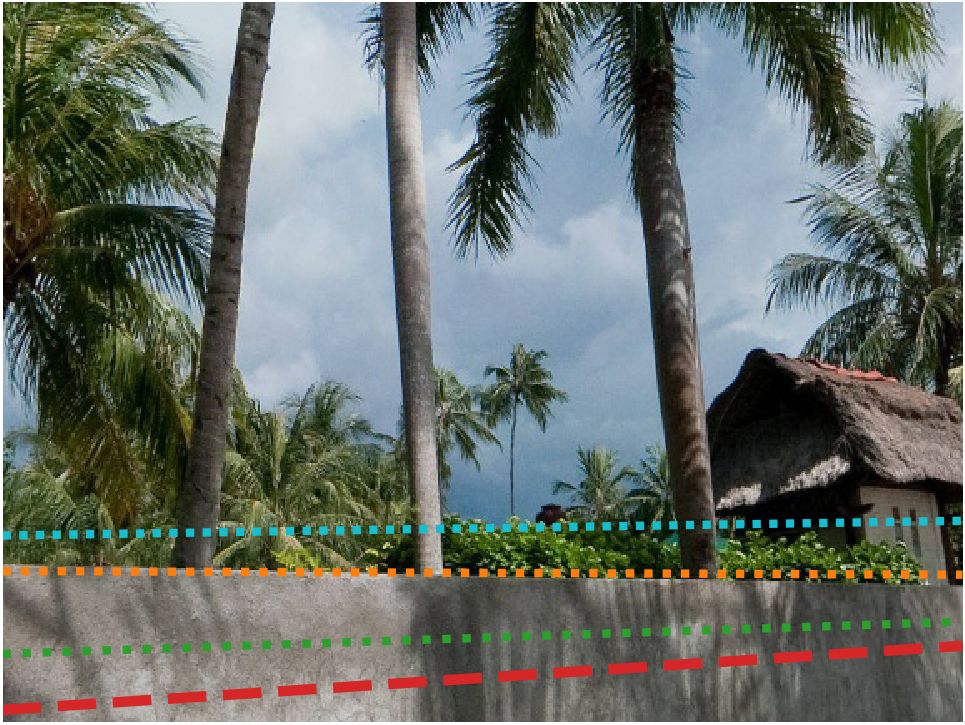} &
\includegraphics[width=\exampleresultswidth\linewidth]{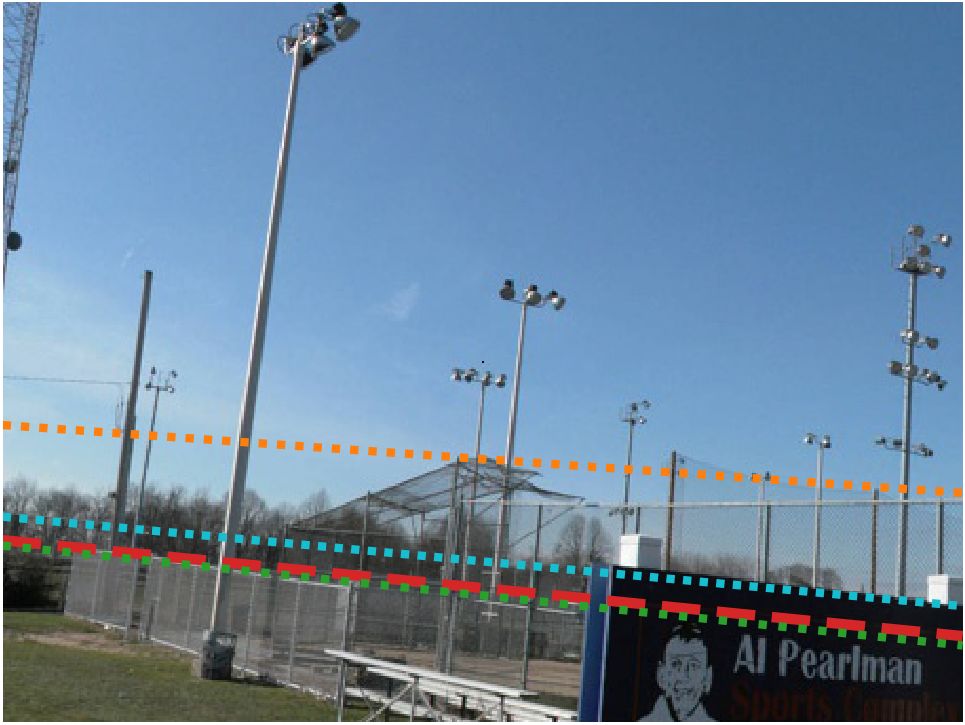} &
\includegraphics[width=\exampleresultswidth\linewidth]{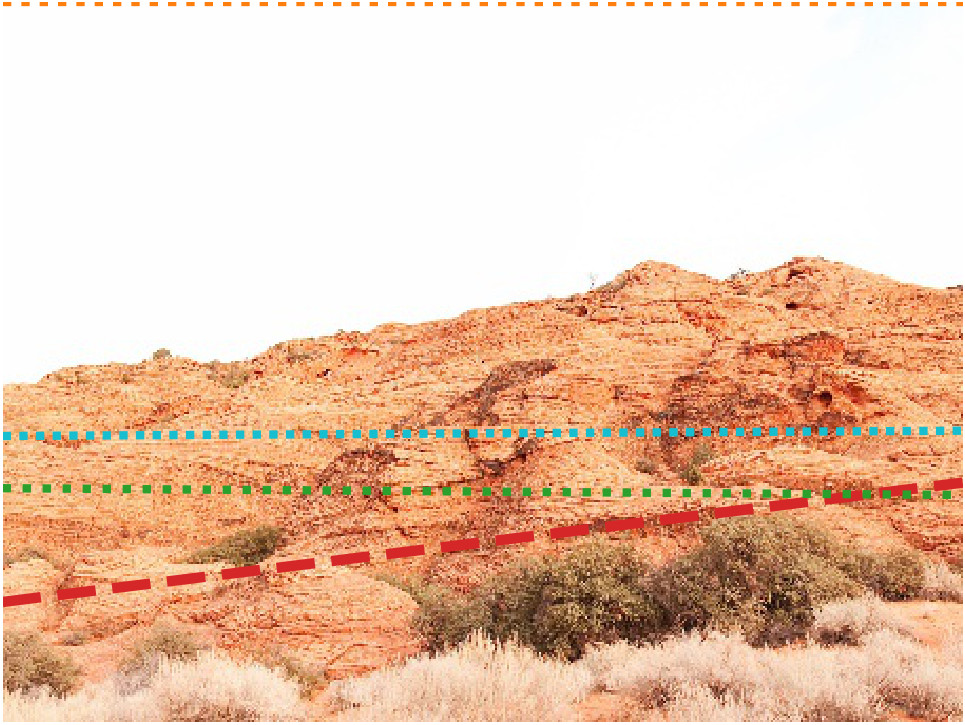} \\
\multicolumn{4}{@{}c@{}}{(a)} & (b) \\
\multicolumn{5}{@{}c@{}}{\includegraphics[trim={0 0.5cm 0.5cm 0},clip,width=0.5\linewidth]{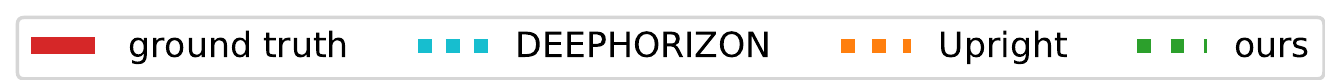}}
\end{tabular}
\egroup
\caption{Example results of horizon line estimation on the SUN360 test set. Note how Upright performs well when sharp human-made objects are present in the scene, whereas deep learned methods are more robust to organic scenes. The last column (b) contains failure cases where either the environment is not well represented in the training set (e.g. auditorium or acoustic panels) or visual cues for vanishing lines are scarce, leading to horizon estimations where humans can potentially be less sensitive to errors. \textbf{More examples available in the supplementary material.}\vspace{-1em}}
\label{fig:method_example_results}
\end{figure*}

%% file: relatedwork.tex

\section{Related Work}







Geometric camera calibration is a widely studied topic that has a significant impact on a variety of applications including metrology~\cite{Criminisi2000}, 3D inference~\cite{Criminisi00,Fouhey2013} and augmented reality, both indoor~\cite{hedau-iccv-09,izadinia-cvpr-17} and outdoor~\cite{hoiem-cvpr-06}. As such, many techniques have been developed to perform precise geometric calibration using a calibration target inserted in the image~\cite{Sturm1999,Zhang2002,Heikkila1997,Chen2004}. For after-the-fact calibration, most work on camera calibration aims to detect specific geometric objects in the image typically present in human-made environments~\cite{Rother2000,Melo2013}. Similarly, PoseNet~\cite{kendall-iccv-15} performs camera relocalization by jointly learning location and orientation. Methods for straightening photographs like Upright~\cite{Lee2014} recover calibration by finding vanishing points. Other work has proposed to take advantage of lighting cues for camera calibration~\cite{Lalonde2010,Workman2014}, circumventing the need to detect vanishing lines. However, these techniques often fail on complex scenes where semantic reasoning is required to discard misleading textures and visual cues. To solve the need for high-level reasoning, deep convolutional neural networks were recently used to estimate field of view~\cite{Workman2015a} and horizon lines~\cite{Workman2016}, bringing camera calibration on single images to a wider variety of scenes.

Understanding the limits of the human visual system has also received significant attention, with studies quantifying color sensitivity~\cite{fairchild2013color}, how reliably we can detect photo manipulations artifacts~\cite{Farid2010} and how people perceive distortion in street-level image-based rendering~\cite{Vangorp2013}. More recently, perceptual studies were performed to assess human appreciation on tasks like super-resolution~\cite{ledig-cvpr-17}, image caption generation~\cite{vinyals-cvpr-15} and video temporal alignment~\cite{papazoglou-accv-16}.

In this work, we go one step further by proposing a CNN-based method estimating jointly field of view and the horizon line and understanding human sensitivity to calibration errors and comparing the features sought by our method to traditional vanishing-lines-based methods.

%% file: representation.tex
\section{Geometric camera model}
\label{sec:camera-model}

We first present the geometric camera model used in this paper. 
Under the pinhole camera model, the pixel coordinates $\mathbf{p}_\mathrm{im}$ of a 3D point $\mathbf{p}_\mathrm{w}$ is given by
\setlength{\abovedisplayskip}{5pt}
\setlength{\belowdisplayskip}{5pt}
\begin{equation}
\mathbf{p}_{\mathrm{im}} = [\lambda u \; \lambda v \; \lambda]^T = \mathbf{K} \left[\mathbf{R} | \mathbf{t}\right] \left[ \mathbf{p}_{\mathrm{w}} | 1 \right]^T
\end{equation}
in homogeneous coordinates, where $\mathbf{K}$ is the camera projection matrix (camera intrinsics), $\mathbf{R}$ and $\mathbf{t}$ are the camera rotation and translation in the world reference frame (camera extrinsics). Simplifying the model further to square pixels, no skew, and image center at the principal point, the projection matrix $\mathbf{K}$ is given by $\mathbf{K} = \mathrm{diag}([f_{\mathrm{px}} \; f_{\mathrm{px}} \; 1])$, where $f_{\mathrm{px}}$ is the focal length in pixels. 

Since the camera parameters are to be estimated from a single image, we first express them as a function of image features. Let us first consider the focal length $f_\mathrm{px}$. Since it has no direct interpretation in the image, we instead estimate the vertical field of view $h_\theta$, a more intuitive measure:
\vspace{-0.9em}
\begin{equation}
h_{\theta} = 2 \arctan \left( \nicefrac{ h }{ 2f{_\mathrm{px}} } \right) \,,
\end{equation}
where $h$ is the image height.

We next consider the rotation matrix $\mathbf{R}$, which can be parameterized by roll $\psi$, pitch $\theta$, and yaw $\varphi$ angles. There exists no natural reference frame to estimate $\varphi$ (left vs right) from an arbitrary image. Therefore, we constrain the rotation to only pitch and roll components, simplifying the extrinsic rotation matrix to $\mathbf{R} = \mathbf{R}_z(\psi) \mathbf{R}_x(\theta)$. We can use the horizon line as an intuitive representation for these angles. We define the horizon line midpoint $b_{\mathrm{p}}$ as the $y$-coordinate of its intersection with the vertical axis in the image and roll $\psi$ with respect to horizontal.

The midpoint $b_{\mathrm{p}}$ can be derived from $\theta$ and $f_{\mathrm{px}}$ as
\begin{equation}
b_{\mathrm{p}} = 2 f_{\mathrm{px}} \tan\theta \,.
\label{eq:horizon_midpoint}
\end{equation}
In this image units representation, the top and bottom of the image have coordinates 1 and $-1$ respectively. 

Throughout our work, camera calibration refers to the vertical field of view $h_{\theta}$, pitch $b_{\mathrm{p}}$ and roll $\psi$ from this simplified geometric camera model.

%% file: method.tex
\section{Image calibration network}
\label{sec:proposed_method}

In this section, we present our CNN architecture for single image calibration  and compare it to the state-of-the-art estimation method.  To train this model, we need a large number of images and their corresponding camera parameters. However, existing datasets either provides relatively accurate field of view~\cite{Wilson2014} or horizon lines~\cite{Workman2016}, but not both simultaneously.
In the following, we discuss how we generate our camera calibration dataset, how the CNN model was trained and what are the probable cues it looks for to perform camera calibration.

\subsection{Dataset}
\label{sec:dataset_generation}

\begin{table}[!t]
\centering
\footnotesize
\begin{tabular}{lll}
\toprule
Parameter & Distribution & Values \\
\midrule
Focal length (mm) & Lognormal & s = 0.8, loc = 14, scale = 17 \\
Horizon (im. height) & Normal & $\mu=0.046, \sigma=0.6$ \\
Roll (rad) & Cauchy & $x_0 = 0$, $\gamma \in \{0.001, 0.1\}$ \\
Aspect ratio & Varying & $\{1{:}1, 5{:}4, 4{:}3, 3{:}2, 16{:}9\}$ \\
\bottomrule
\end{tabular}
\vspace{1em}
\caption{Sampling of camera parameters used to generate the dataset for the human sensitivity study.}
\label{tab:parameters-sampling}
\end{table}

Our goal is to train a deep network to estimate the camera roll, pitch, and field of view from a single image. To obtain images and their ground truth camera parameters, we take inspiration from \cite{Workman2016,Hold-Geoffroy2017} and leverage the SUN360 database~\cite{Xiao2012}, which contains a large number of $360^\circ$ panoramas. We extract 7 rectified images from each panorama using a standard pinhole camera model of random parameters. To obtain reasonable camera parameters, the sampling strategies indicated in table~\ref{tab:parameters-sampling} were employed. Note that for the camera roll, two different Cauchy distributions are sampled with 0.33 and 0.66 probability respectively. This was done to model the fact that many photos typically have a roll close to 0. The aspect ratios were chosen by sampling Flickr and ImageNet images. Note that a larger probability (0.6) was given to the $4{:}3$ aspect ratio as it is the most common. The other aspect ratios are given a probability of $0.1$. We resize the extracted images to $224\times224$ to fit the neural network input size. This results in a dataset of 399,728 pairs of photos and their corresponding camera calibration which we split into a training set of 389,760 pairs, a validation set of 9,078 pairs and a test set of 890 pairs. Special care was taken to ensure no panorama used in the training set was present during validation or test.

As in~\cite{Workman2016}, we use the slope/offset ($\psi,\rho$) representation, where the horizon line is parameterized by the roll $\psi$ and its perpendicular distance from the center of the image $\rho$.

\subsection{Architecture}

We adopt a DenseNet~\cite{Huang2016} model pretrained on ImageNet~\cite{Russakovsky2015} on which the last layer is replaced with three separate heads: one for estimating the horizon angle $\psi$, a second one to estimate the horizon's distance to the center of the image $\rho$ and a third one to estimate the vertical field of view of the image $h_{\theta}$.
All output layers use the softmax activation function to output a probability distribution by discretizing their respective parameter into 256 bins. This type of representation was also used in~\cite{Workman2016}.
We adopt a range of $\left[ -\nicefrac{\pi}{2}, \nicefrac{\pi}{2} \right]$ for slope and $\left[ -1.6, 1.6\right]$ for offset. For both parameters, we use smaller bins around 0 for finer estimations around those values. Bin width follow an inverse normal distribution with $\mu=0,\, \sigma=\left\{ 0.5, 1 \right\}$ for slope and offset, respectively. For field of view, we use bins uniformly distributed over a range of $\left[0.2, 1.8\right]$ radians. We sum the Kullback-Leibler divergence of the three heads and use it as loss to train the model, which we minimize using stochastic gradient descent with the Adam optimizer~\cite{Kingma2015} with an initial learning rate of $\eta = 0.001$ and a learning rate decay of $ \alpha = 0.0002$. Training is performed on mini-batches of 42 images. Convergence is observed through early stopping typically after 9-10 epochs.

\begin{figure}
\centering
\includegraphics[width=\linewidth]{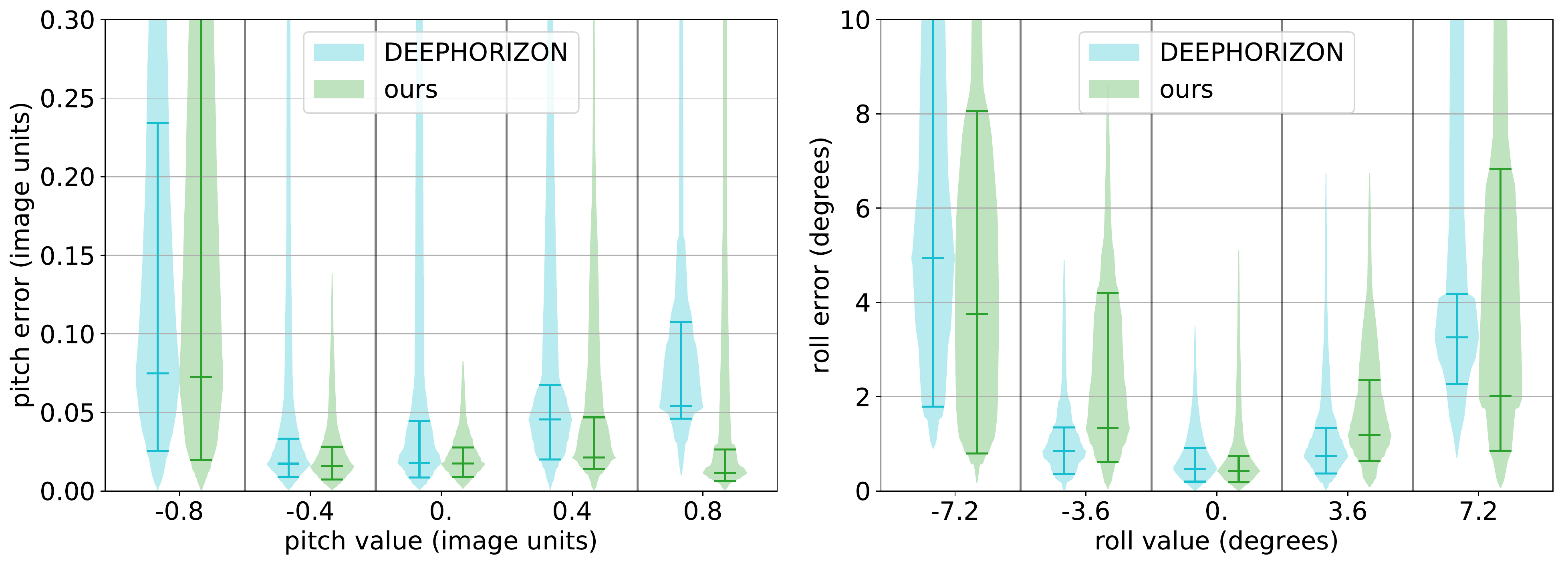} \\
\includegraphics[width=\linewidth]{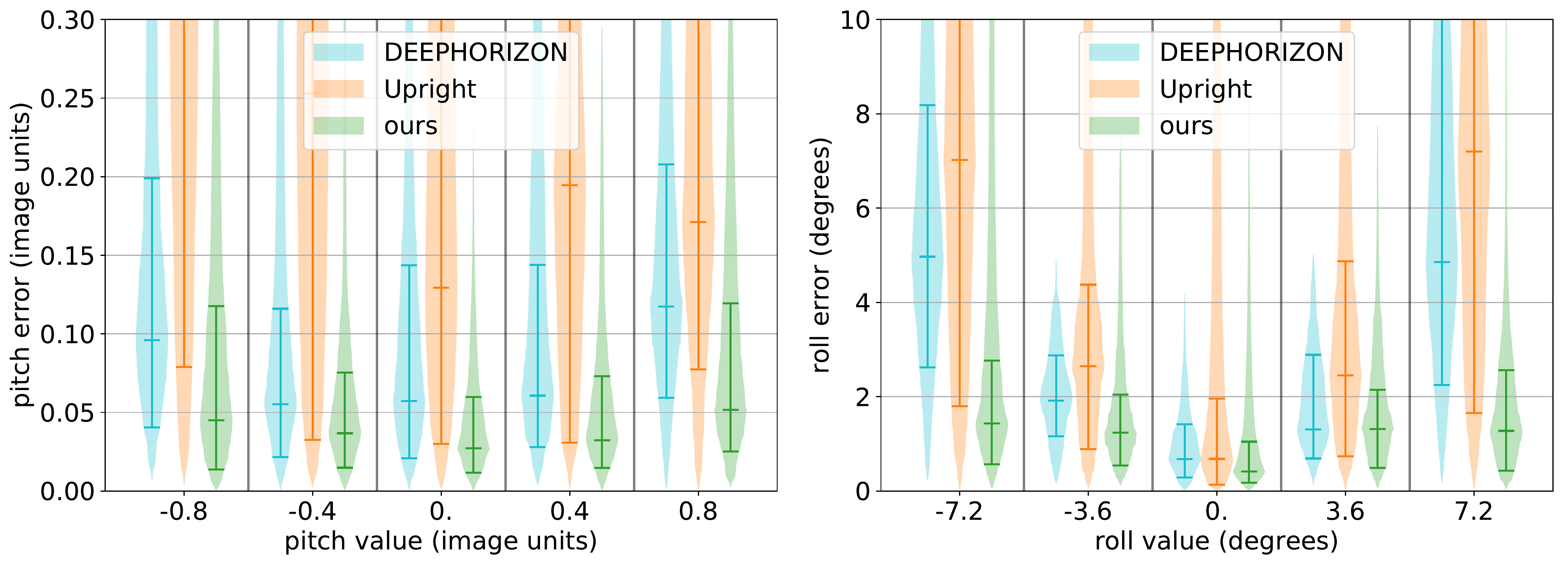}
\caption{Pitch (left) and roll (right) estimation performance on the HLW dataset (top) and our SUN360 test set (bottom). Negative pitch denotes a camera pointing up. Results are displayed as "box-percentile plots"~\cite{esty-jss-03}, where the column envelope represents the percentile and the horizontal bars represents the first quartile, median and third quartile. Estimation errors (y-axis) are grouped into bins according to the parameter value (x-axis).}
\label{fig:method_pitch_roll_performance}
\end{figure}

\begin{figure}
\centering
\includegraphics[width=\linewidth]{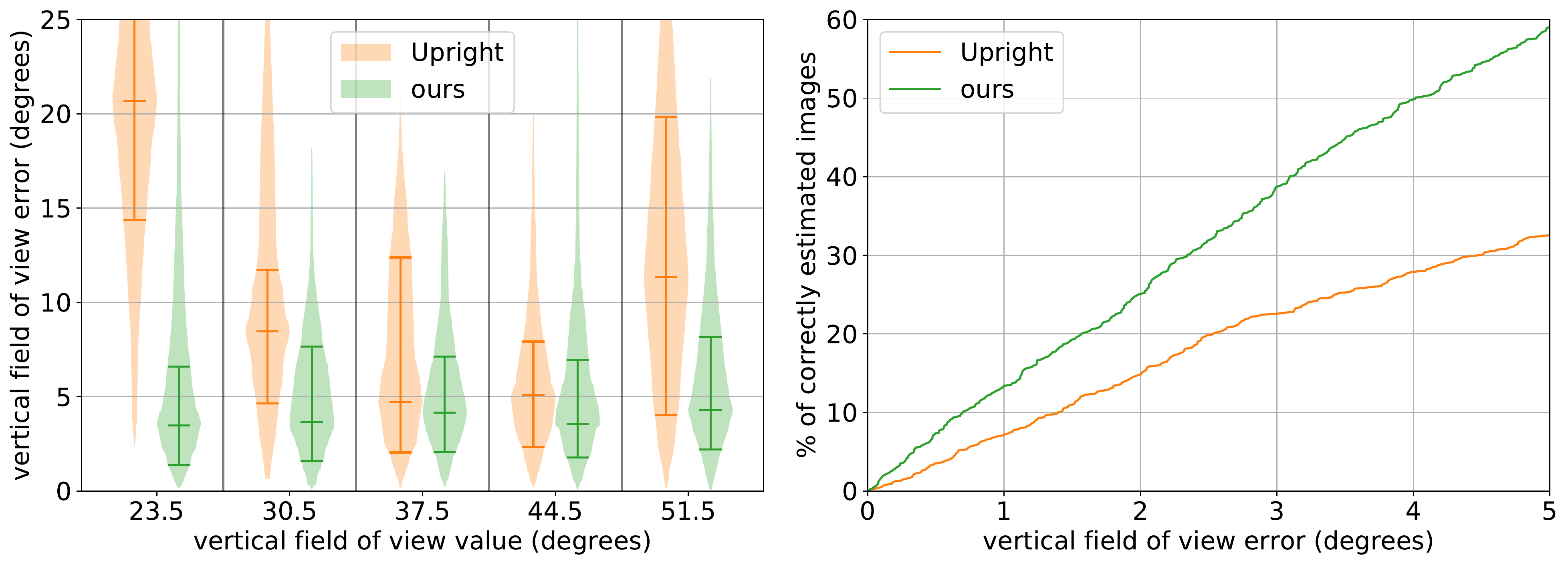}
\caption{Vertical field of view estimation performance on our SUN360 test set displayed as a "box-percentile plot" (left) and a cumulative distribution function (right). See fig.~\ref{fig:method_pitch_roll_performance} for an explanation of the box-percentile plot.
}
\label{fig:method_vfov_performance}
\vspace{-1em}
\end{figure}


\section{Evaluation}

We report quantitative horizon line estimation performance of Upright~\cite{Lee2014}, DEEPHORIZON~\cite{Workman2016} (as trained by the authors) and our method (trained on SUN360) on two different datasets, the Horizon Lines in the Wild (HLW) dataset~\cite{Workman2016} and our SUN360 test set in fig.~\ref{fig:method_pitch_roll_performance}. We observe that aside from large roll errors on HLW, our method outperforms the state-of-the-art in most cases. Upright fails to converge on many cases since multiple images in our SUN360 test set do not contain edges on which the technique relies on. Qualitative results are shown in fig.~\ref{fig:method_example_results}.



Fig.~\ref{fig:method_vfov_performance} shows quantitative field of view estimation accuracy on our SUN360 test set.
Our method significantly outperforms Upright~\cite{Lee2014} across the entire range of parameters.

\newcommand{\sgbpwidth}{0.2}
\begin{figure*}
\centering
\bgroup
\def\arraystretch{0}
\begin{tabular}{@{}c@{}c@{}c@{}c@{}c@{}}
\includegraphics[width=\sgbpwidth\linewidth]{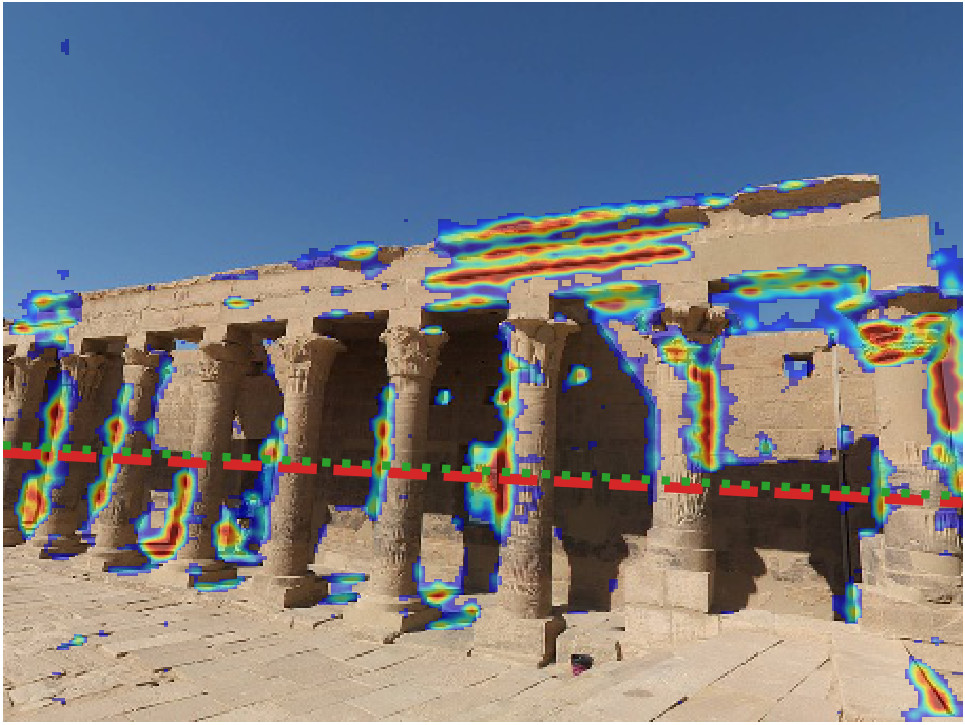} &
\includegraphics[width=\sgbpwidth\linewidth]{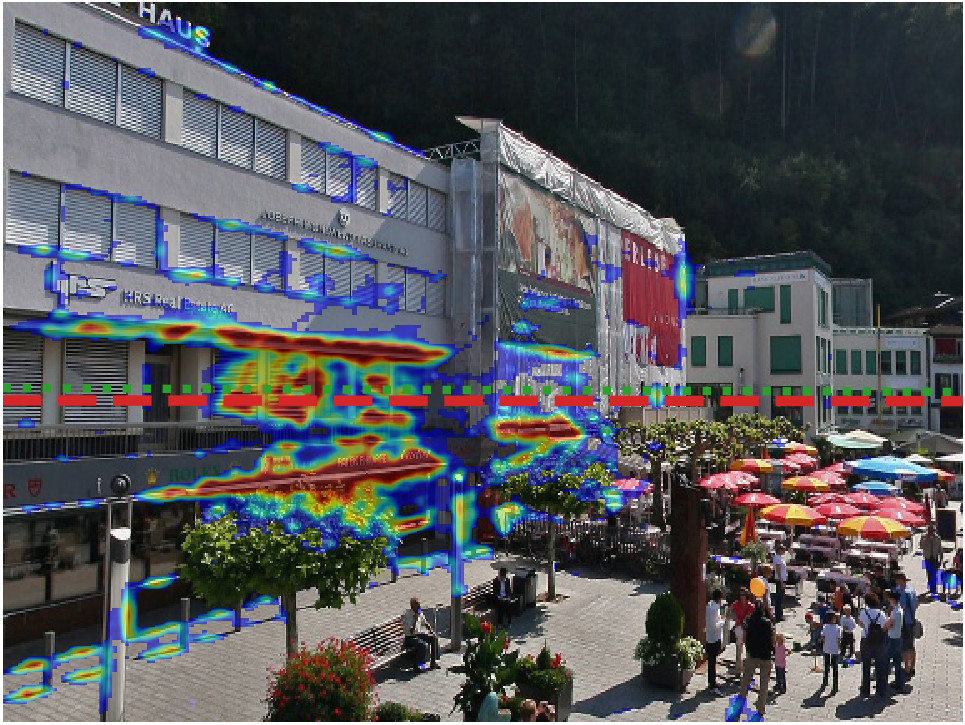} &
\includegraphics[width=\sgbpwidth\linewidth]{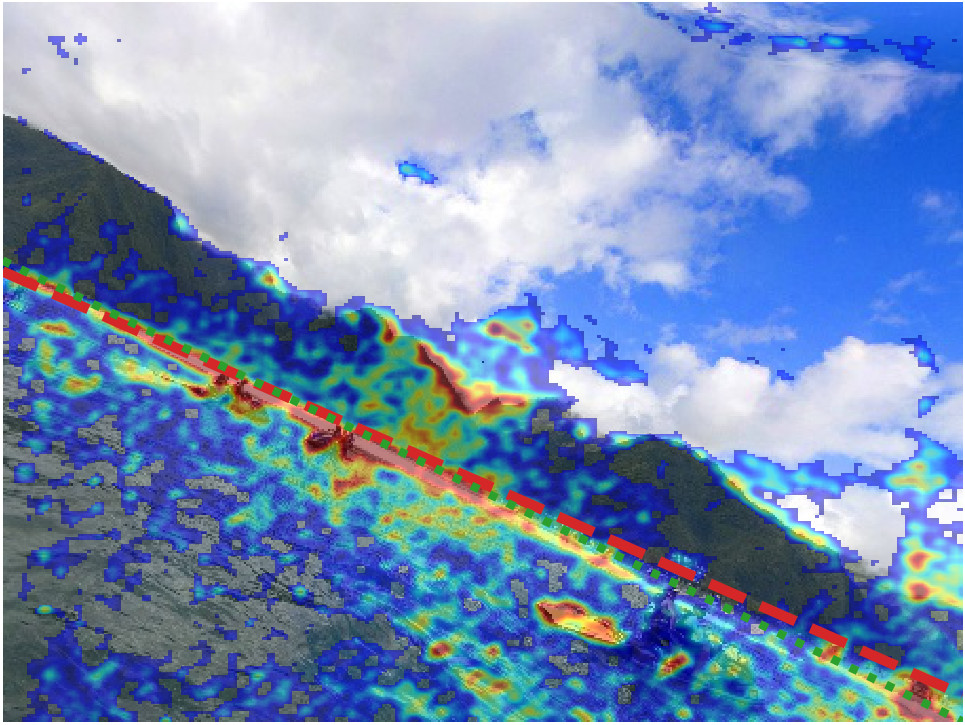} &
\includegraphics[width=\sgbpwidth\linewidth]{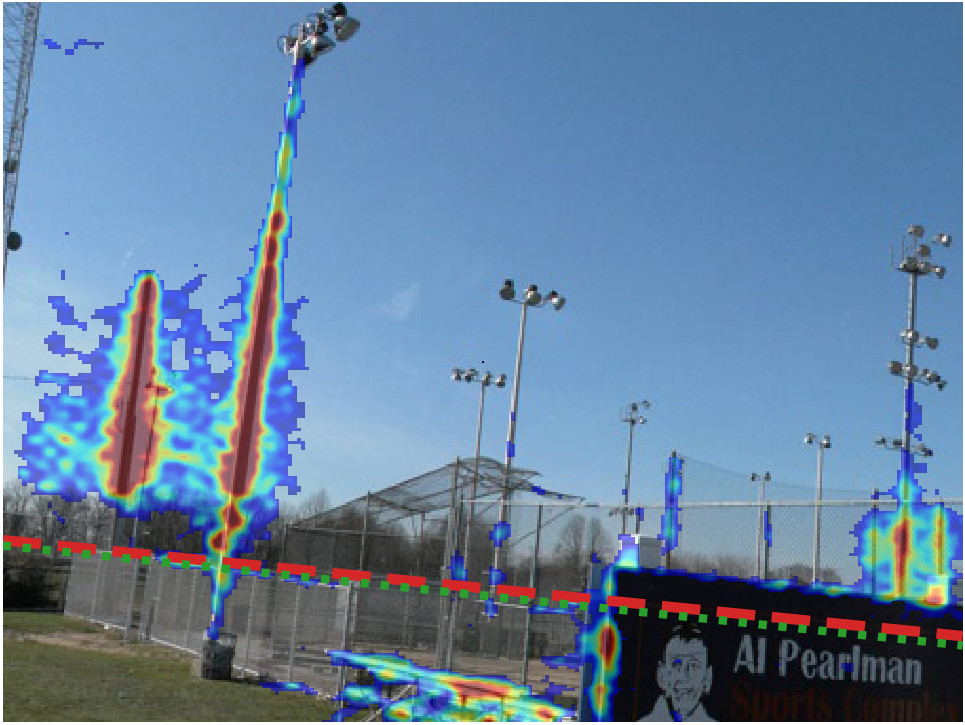} &
\includegraphics[width=\sgbpwidth\linewidth]{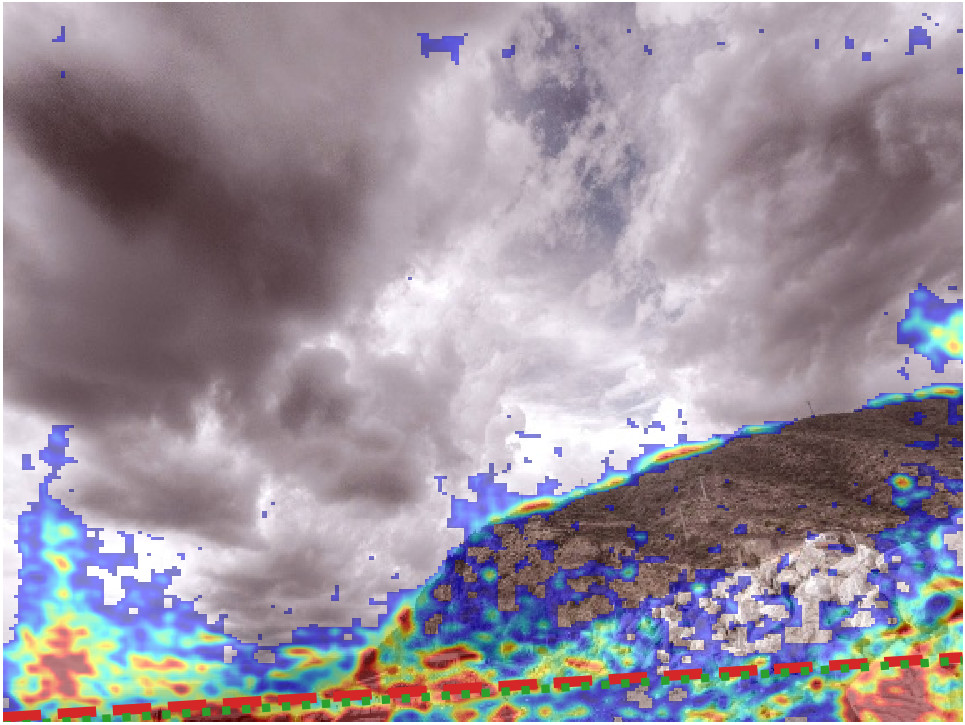} \\
\includegraphics[width=\sgbpwidth\linewidth]{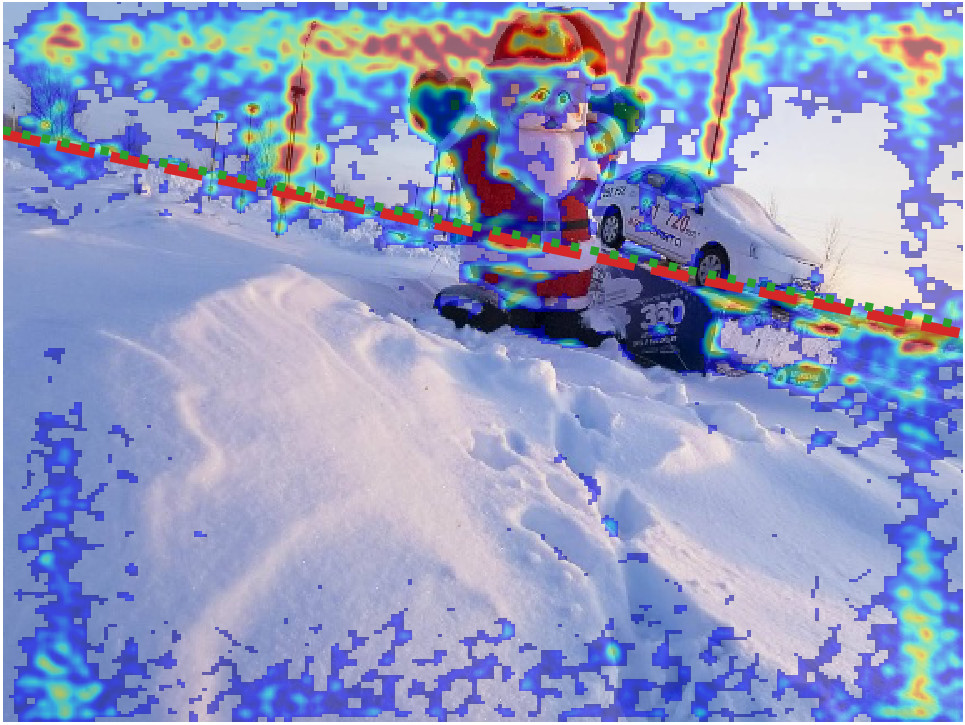} &
\includegraphics[width=\sgbpwidth\linewidth]{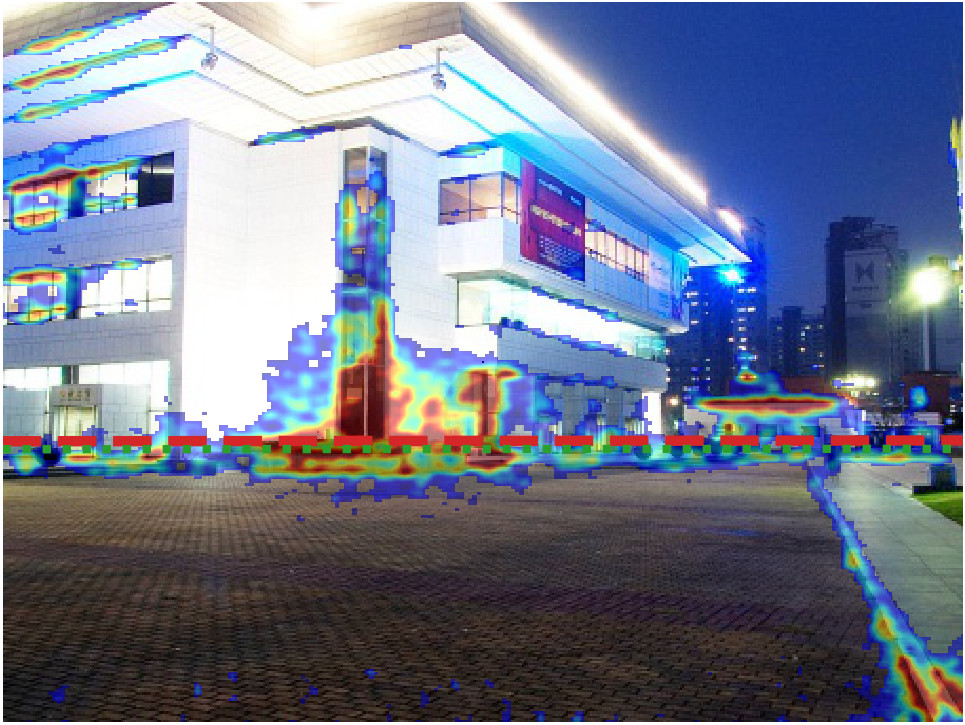} &
\includegraphics[width=\sgbpwidth\linewidth]{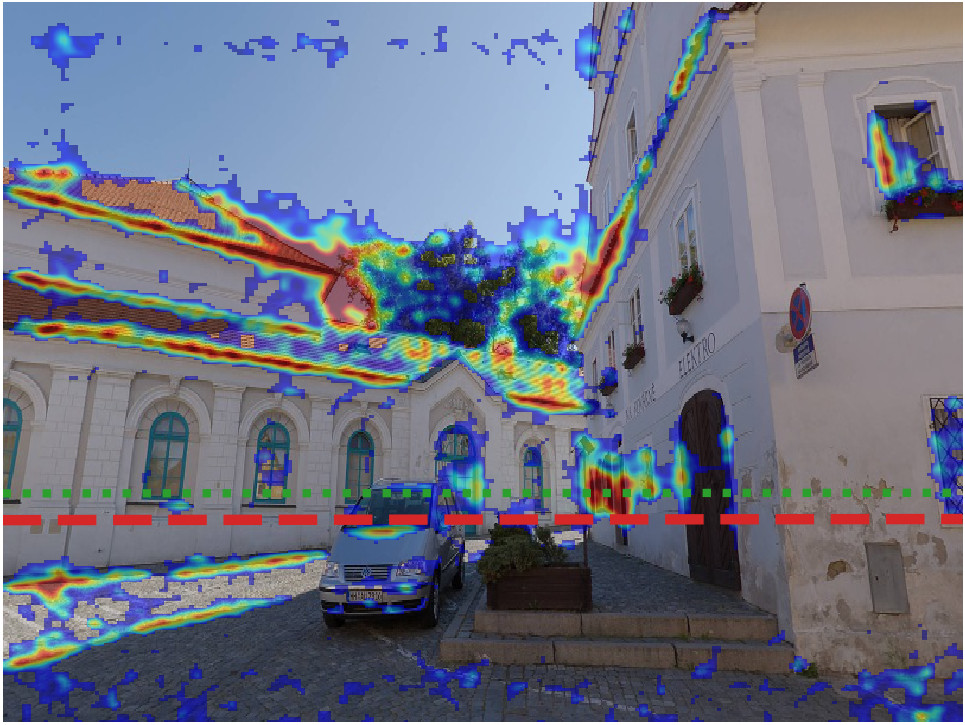} &
\includegraphics[width=\sgbpwidth\linewidth]{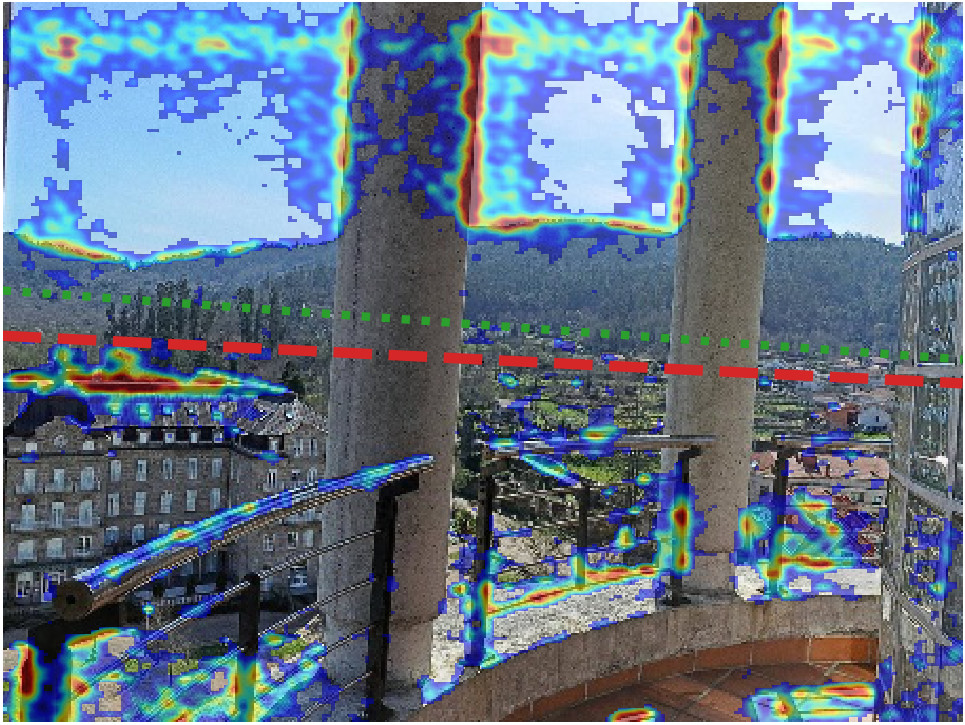} &
\includegraphics[width=\sgbpwidth\linewidth]{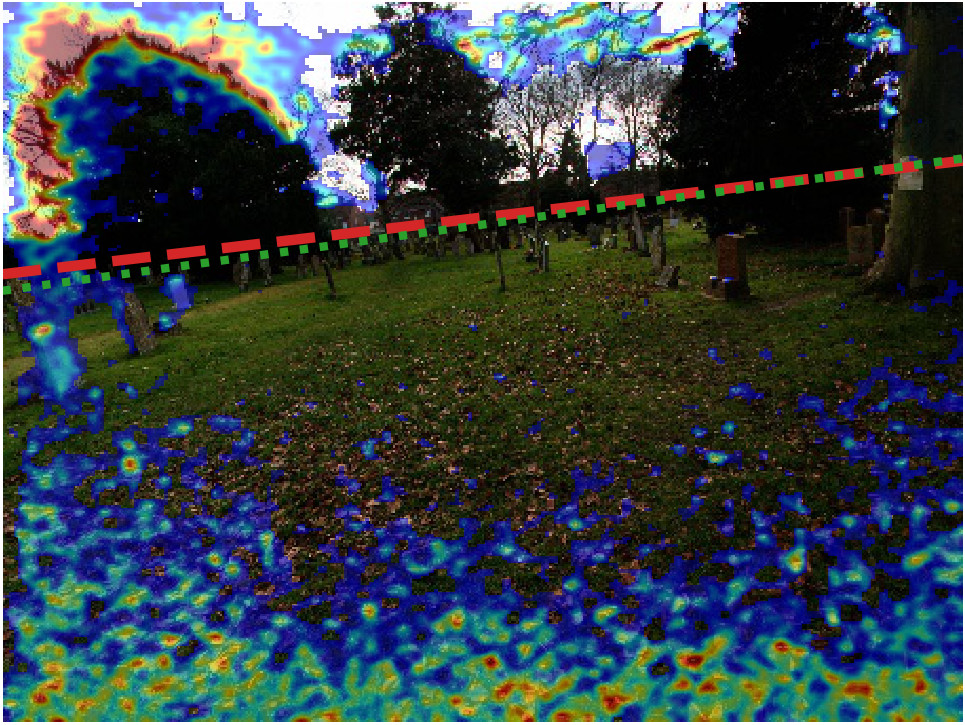} \\
\multicolumn{5}{@{}c@{}}{\includegraphics[trim={0 0.5cm 0.5cm 0},clip,width=0.3\linewidth]{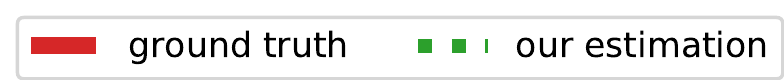}}
\end{tabular}
\egroup
\def\arraystretch{0.5} 
\caption{\textbf{Analysis of the neural network focus.} The result of smoothed guided backpropagation is displayed as a jet overlay. When present, edges corresponding to important vanishing lines are highlighted while other edges are discarded. When no clear horizontal vanishing lines are detected, the neural network seems to look for the boundaries of either sky or land textures while dismissing the clouds or objects like trees, probably hinting bounds on horizon location in the image. \textbf{More examples available in the supplementary material.}}
\label{fig:nn_analysis_smoothed-guided-back-propagation}
\end{figure*}

\paragraph{Feature analysis} We use guided backpropagation~\cite{Springenberg2015} to understand the image features our CNN-based method focuses on to perform its estimation. We use the smoothgrad~\cite{Smilkov2017} version of guided backpropagation (SGB) to obtain a more stable analysis. Qualitative results are shown in fig.~\ref{fig:nn_analysis_smoothed-guided-back-propagation}. Note how edges representing vanishing lines are highlighted by SGB in accordance to the features used by geometry-based approaches such as Upright. Sharp edges that are not useful for horizon estimation, such as clouds or organic objects,  are not taken into account. As such, we believe this focus map could help geometric-based approaches select the appropriate edges for geometric calibration estimation. Furthermore, when no clear vanishing line is detected in the image, the CNN model tends to focus on boundaries between sky and land, as the horizon typically lies on or below this boundary.

%% file: pstudy.tex
\section{Human perception of calibration}
\label{sec:human_sensitivity_analysis}

Given the fundamental role it plays in the context of geometric scene reconstruction, single image camera calibration has been studied extensively. Most of this work attempts to \emph{exactly} recover the camera calibration. However, it has been observed that the human visual system is forgiving of inconsistencies in perspective for single image applications like virtual object compositing~\cite{karsch2014automatic}. In this work, we aim to understand what the bounds of these human tolerances are. Knowing these can allow us to design a) camera calibration methods that match human performance, and b) geometric editing applications like object compositing that can ``fool'' human observers.

In particular, we aim to understand the sensitivity of humans to camera calibration errors in the context of virtual object insertion by running a large-scale user study on Amazon Mechanical Turk. In the following, we discuss how we generated the dataset necessary for the study, how the user study was performed, and provide a detailed analysis of the obtained results.

\subsection{Dataset generation}
\label{sec:dataset_organization}

To understand human perception of geometric camera calibration errors, we show pairs of images containing a virtual object to multiple users, one aligned according to the ground truth camera calibration and the second with some parameter(s) distorted to some extent. To generate this dataset, we use the same process as described in~\ref{sec:dataset_generation} to obtain images with their ground truth calibration. We randomly selected 530 panoramas (79 indoor and 451 outdoor) from this dataset, resulting in 10,638 images with realistic camera parameters. 

An insertion point on the ground was manually selected for each image. The Cycles renderer was used to realistically insert virtual objects at that point in the images. The ground plane was set at $y=0$ using the shadow catcher feature. The virtual camera was placed at a height of 1.6m. Automatic lighting estimates were obtained by leveraging recent work in single image lighting estimation for the outdoor~\cite{Hold-Geoffroy2017} and indoor~\cite{Gardner2017} cases. 

Two renders are generated for each image. The first one is obtained by setting the parameters of the virtual (rendering) camera to the ground truth parameters of the background image. The second render is obtained by distorting the virtual camera parameters, yielding a virtual object that does not have the same camera parameters as the background. For this second render, either pitch, roll, field of view or a combination of these parameters were modified. The parameters were altered by randomly adding or subtracting values sampled from a uniform distribution in $\left[ 1, 30 \right]\degree$ for pitch, $ \left[ 0.5, 20 \right]\degree$ for roll and $\left[ 5, 55\right]\degree$ for field of view. In the distorted renders, the object was moved and scaled in order to appear at the same location and have the same size in the image as the ground truth render. This step is needed as apparent size is a function of field of view. The virtual objects were laid vertically on the ground plane and were randomly rotated about their vertical axis.

To limit biases caused by the virtual object inserted, we inserted 8 different virtual objects on each image. These objects include simple geometric primitives (sphere, cone), real objects with clear vertical directions (toy rocket, metal barrel, the Eiffel tower) and objects with a somewhat organic shape (the Stanford bunny and a horse statue). Several examples of images generated using this process are shown in fig.~\ref{fig:pstudy_sensitivity_per_parameter} and in the supplementary material. 

\subsection{Perceptual evaluation}
\label{sec:hsa_perceptual_evaluation}

We use Mechanical Turk to perform a perceptual study where workers were shown two renders of the same object on the same background image: one with ground truth parameters, another one with distorted parameters (see \ref{sec:dataset_organization}). Workers were asked to select the image where the object orientation looks better. To help them focus on camera parameters, they were specifically instructed to ignore the color, texture, shadows or lighting on or around the object. This form of a forced choice A/B test allows us to isolate the effect of the camera calibration and discard any potential issues caused by the way we create these composites (e.g., non-realistic objects, inaccurate lighting estimation, etc.)

In total, 376 workers provided 145,720 submissions, from which 124,740 were accepted, leading to 11 different users annotating each image on average. 4319 of those submissions had a single distorted parameter, while the remaining 5947 had two or more distorted parameters. To ensure quality work, sentinels~\cite{Gingold2012} were inserted throughout the experiment. These sentinels are manually validated images with obviously distorted calibration. Workers were presented batches of 20 images at a time which contained 2 sentinels. 9 workers were blocked from repeatedly selecting the sentinels. An additional 520 annotations were rejected for inattentive workers failing some sentinels over a small time lapse. The median time spent on a single pair of images was 4 seconds.

\subsection{Study results}

In this section, we report on the analysis performed on the user study results. We analyze the impact of several aspects: the virtual objects rendered in the images, the background images, the ground truth camera parameters, and the joint space of error in parameters with themselves.

\paragraph{Virtual object}

We found no statistical difference in the results obtained across the virtual objects, except for two: the sphere and the Stanford bunny. For these two objects, the participants in the study were unable to identify the ground truth image except for significant field of view variations (at least $30^\circ$). The sphere being rotationally symmetric, it is unsurprising that it does not serve as a good barometer for determining errors in object insertion. We theorize that the bunny, with its round shape, shares similarities to the sphere in that respect. 

We also found no statistical difference in the results when analyzing the impact of the object size, computed as the relative height of the object with respect to the image. Participants showed similar accuracy regardless of object size (ranging from 10\% to 85\% image height in our dataset).

\paragraph{Background image}

We manually labeled the images into either outdoor or indoors, and found no statistical difference between the two subsets. We also tried more finely-grained labels (e.g. built vs natural) but did not find any interesting trend there either. Finally, we analyzed the impact of the mean image brightness, hypothesizing that a mismatch in camera parameters may be more easily observable in brightly-lit images. We found no statistical differences between images of different mean intensities.

\begin{figure}
\centering
\includegraphics[width=\linewidth]{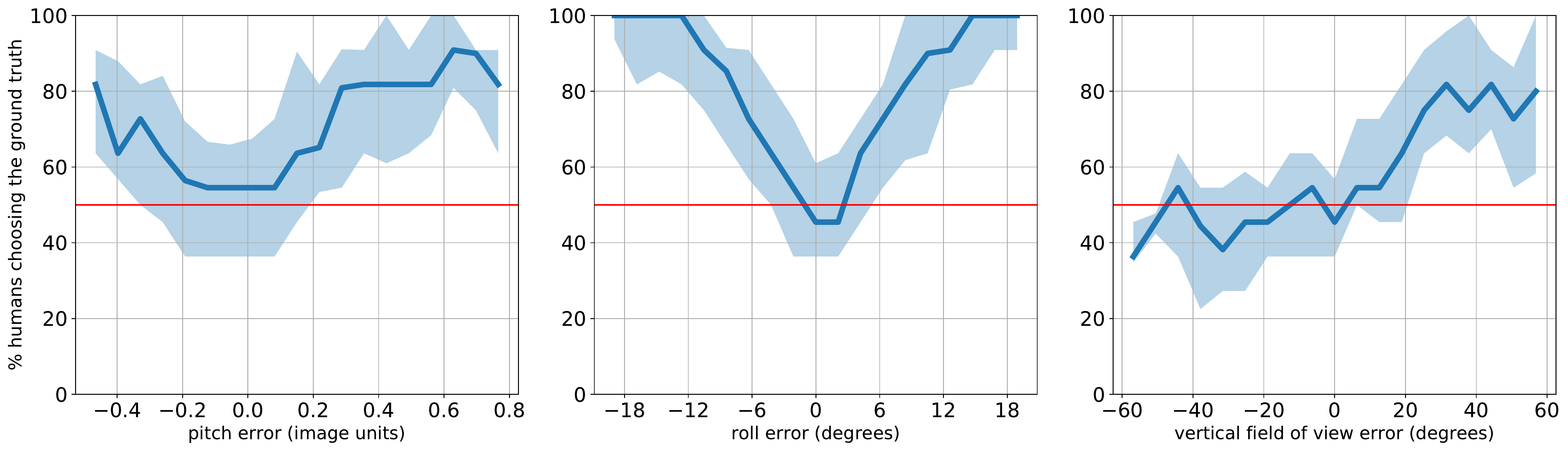} \\
\vspace{-0.1cm}
\begin{tabular}{p{0.3\linewidth}p{0.3\linewidth}p{0.3\linewidth}}
\hspace{1.15cm}(a) & \hspace{1cm}(b) & \hspace{0.85cm}(c)
\end{tabular}
\caption{Human sensitivity to errors in (a) pitch, (b) roll and (c) field of view. We use the percentage of users choosing the image with an object inserted with the ground truth calibration as a measure of human sensitivity. 50\% represents confusion, meaning users are as likely to choose an insertion with a distorted calibration than the ground truth, and 100\% means that all humans could detect the ground truth image. The line represents the median of the percentage of people who pick the ground truth human for each image. The light shaded regions shows the first and third quartile.}
\label{fig:pstudy_overall_sensitivity}
\end{figure}

\paragraph{Error in camera parameters} We evaluate the sensitivity of humans to errors in camera parameters as a function of the errors in each parameter independently and illustrate the results in fig.~\ref{fig:pstudy_overall_sensitivity}. To generate this curve, we compute the percentage of times the study participants preferred the ground truth over the distorted image, and compute the median and percentiles across all images (and different virtual objects) that share the same amount of distortion. The higher the percentage in the $y$ axis, the more humans are prone to detect errors in this scenario. Conversely, 50\% indicates perfect confusion: participants are unable to distinguish between the ground truth and the distorted version. 

First, we note that when the error in camera parameters is close to 0, confusion nears 50\%, which is expected. What is interesting is how quickly sensitivity rises when increasing the error. We note a large tolerance to negative errors in field of view (fig.~\ref{fig:pstudy_overall_sensitivity}-c). Large positive errors (right side of the plot) translate to rendering an object with a field of view that is larger than that of the background. This results in increased perspective effects on the object, which tend to be visible. On the other hand, negative errors indicate that the perspective effect is \emph{not} as pronounced on the object as it should be with respect to the background image. In this scenario, participants seemed to have been unable to differentiate between the ground truth and the distorted object. For field of view, a range of $15\degree$ over and up to $50\degree$ under the ground truth value went unnoticed to the users.

Participants could tolerate an error in pitch up to 0.2 in rescaled image units (see sec.~\ref{sec:camera-model}), but beyond this threshold, users started to distinguish the distortions (fig.~\ref{fig:pstudy_overall_sensitivity}-a). The high sensitivity to roll errors is most prominent (fig.~\ref{fig:pstudy_overall_sensitivity}-b), where errors of $12\degree$ and more are almost systematically detected and only a small range of approximately $\pm2.5\degree$ roll error go unnoticed.

\begin{figure*}
\centering
\includegraphics[width=\linewidth]{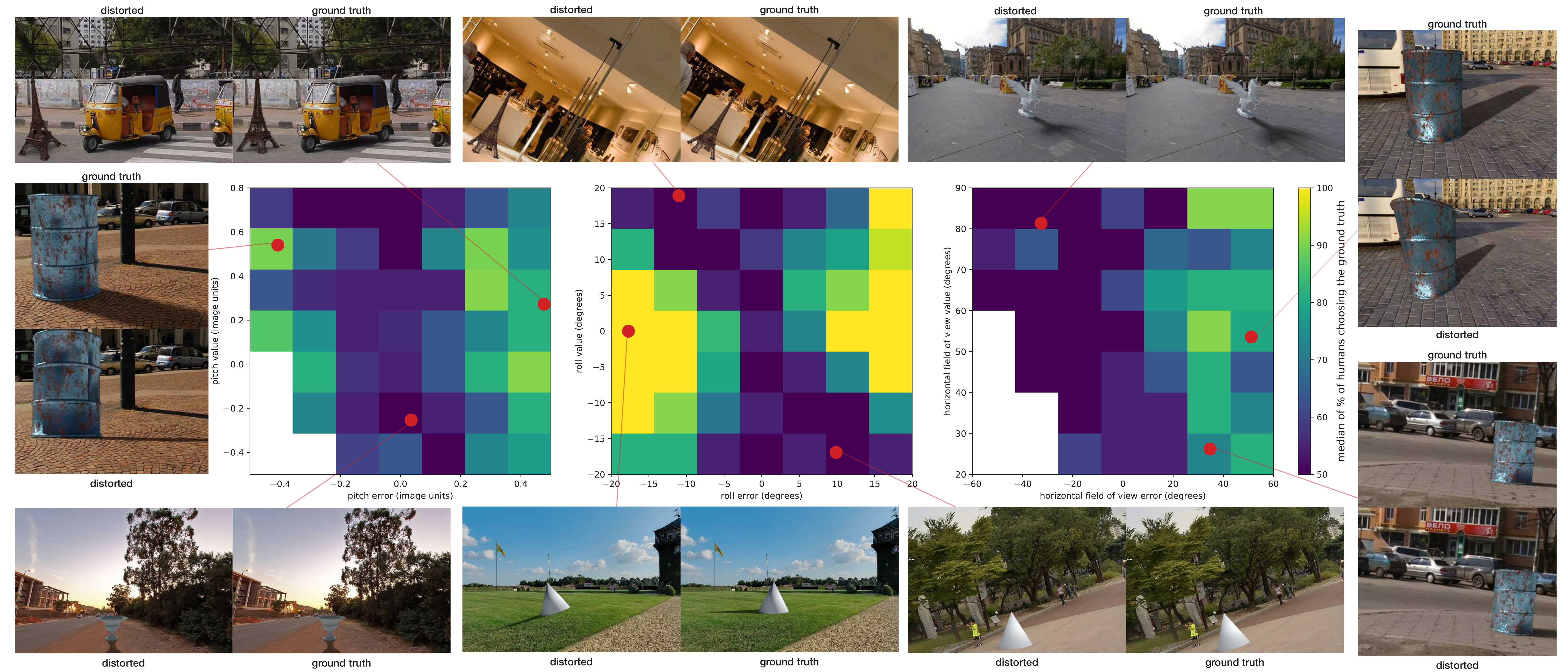}
\caption{Human sensitivity to errors in pitch (left), roll (center) and field of view (right) as a function of individual parameter values, along with examples of image pairs shown to the users. We bin the percentage of people choosing the ground truth per image of the user study. The colors in the plot represents the median over all values in each bin. Note the strong relation between the roll value and its human sensitivity to error. Some combinations of parameters and errors makes it impossible to perform insertion leading to missing values in the figure (e.g., the ground is not visible anymore in the image (bottom left in pitch) or the resulting field of view would be negative (bottom left in field of view)). \textbf{See the supplementary material for more analysis}, including joint modeling of distortions on multiple parameters.}
\label{fig:pstudy_sensitivity_per_parameter}
\end{figure*}

\paragraph{Joint space of error in camera parameters and absolute parameter value}

Evaluating sensitivity to errors in each parameter is interesting, but does not tell the whole story. Are there regions in parameter space where combinations of errors are more noticeable? To evaluate this, we plot the 2D space of errors and absolute parameter value in fig.~\ref{fig:pstudy_sensitivity_per_parameter}.

First, our results suggest that human sensitivity to pitch error does not correlate strongly with horizon position in the image. Similarly, sensitivity to errors in field of view seems to be constant across all fields of view used in the study. However, the camera roll appears to have an influence on our perception of roll error: images with high roll (in either direction) allow more room for roll estimation errors.

\paragraph{}Please see the supplementary material for additional analysis, including joint modeling of distortions on multiple parameters.

\newcommand{\retrievalwidth}{0.12}
\begin{figure*}[!ht]
\centering
\footnotesize
\setlength{\tabcolsep}{3pt}
\begin{tabular}{c|cccc}
\includegraphics[height=\retrievalwidth\linewidth]{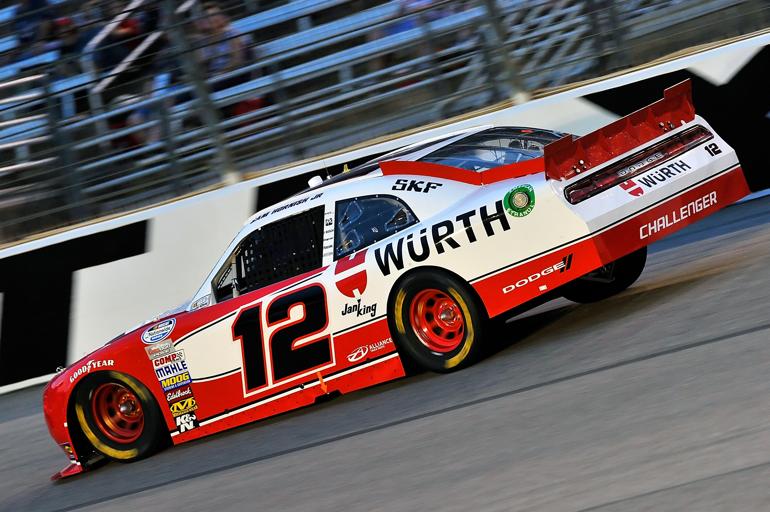} &
\includegraphics[height=\retrievalwidth\linewidth]{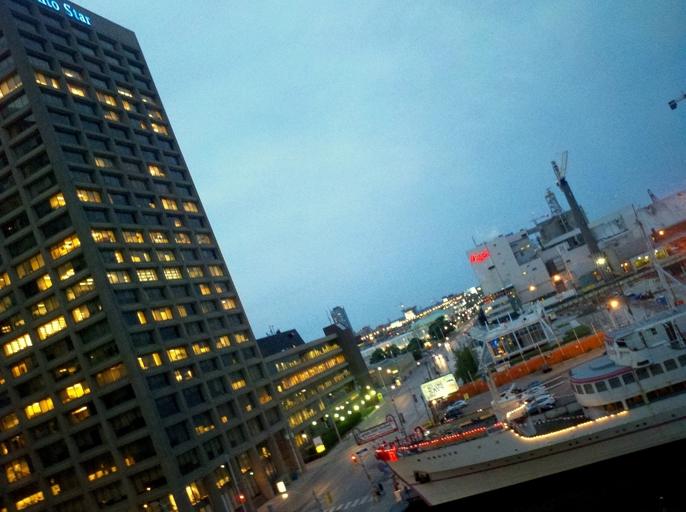} &
\includegraphics[height=\retrievalwidth\linewidth]{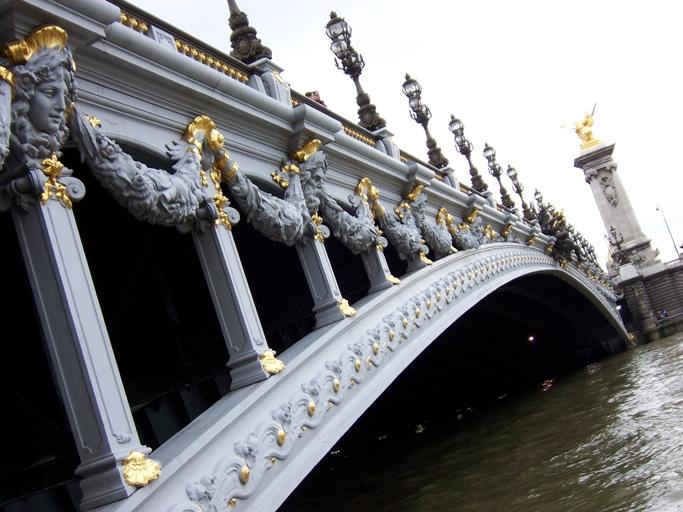} &
\includegraphics[height=\retrievalwidth\linewidth]{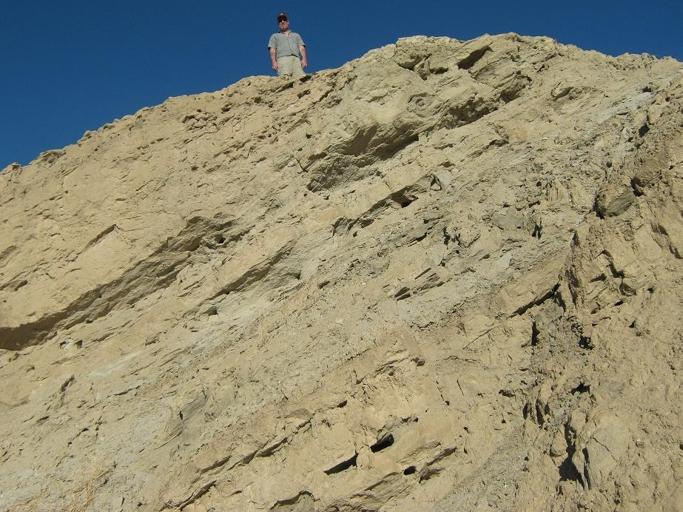} &
\includegraphics[height=\retrievalwidth\linewidth]{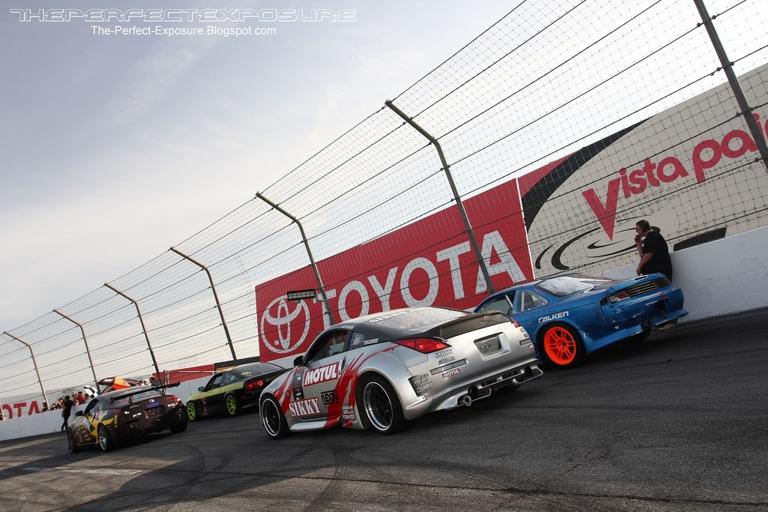} \\
\includegraphics[height=\retrievalwidth\linewidth]{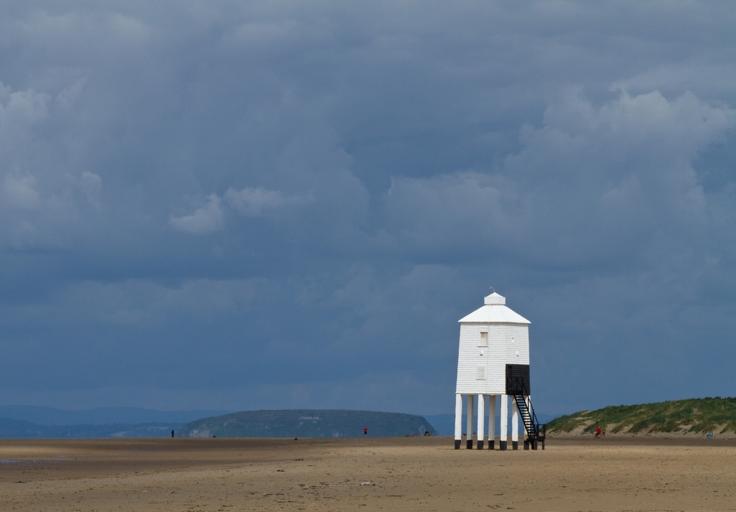} &
\includegraphics[height=\retrievalwidth\linewidth]{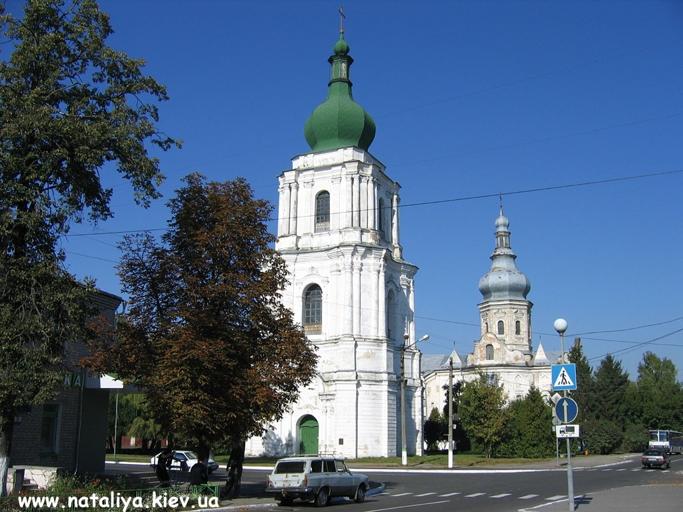} &
\includegraphics[height=\retrievalwidth\linewidth]{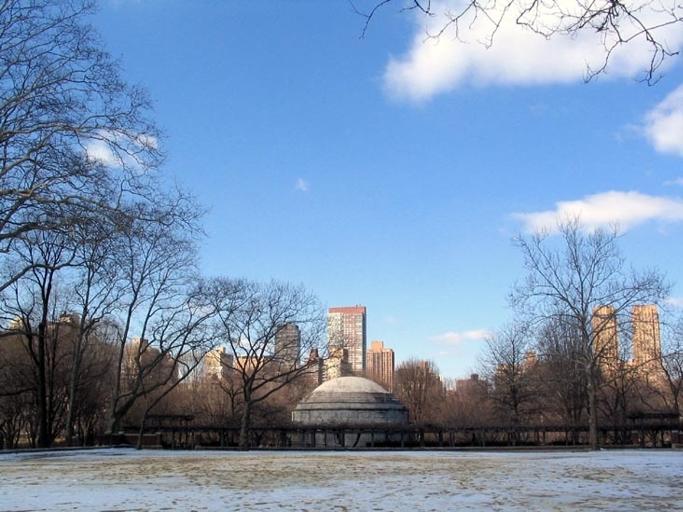} &
\includegraphics[height=\retrievalwidth\linewidth]{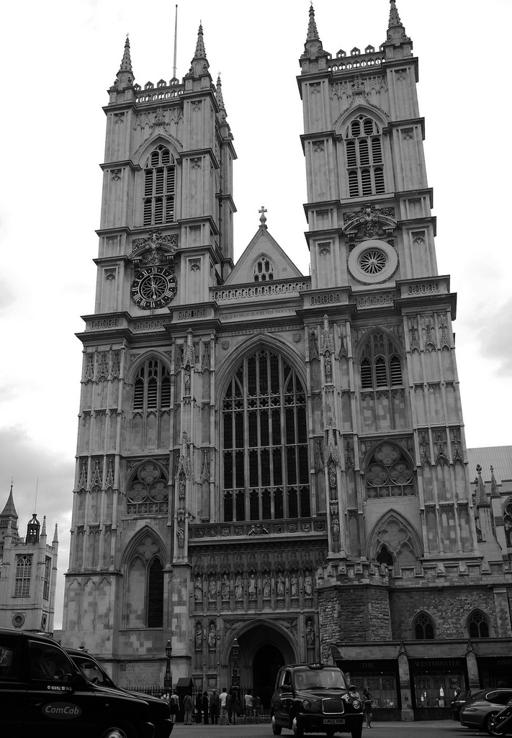} &
\includegraphics[height=\retrievalwidth\linewidth]{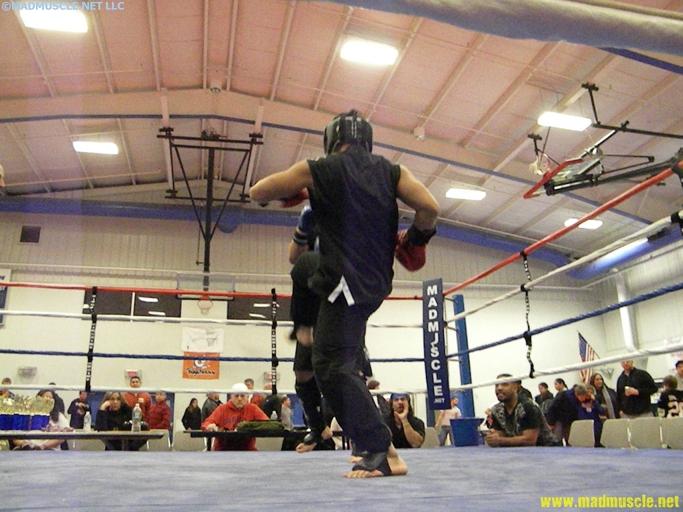} \\
Query & NN1 & NN2 & NN3 & NN4
\end{tabular}
\vspace{1em}
\caption{Examples of image retrieval by horizon location on Places2. The horizon line is estimated using our method from the query image, and used to find closest matches in a 10k random subset from Places2. The top-4 matches are shown on the right.}
\label{fig:applications_retrieval}
\end{figure*}

%% file: method_evaluation_userstudy.tex
\subsection{CNN evaluation on human perception}
\label{sec:cnn-evaluation-perception}

\begin{figure}
\centering
\includegraphics[width=\linewidth]{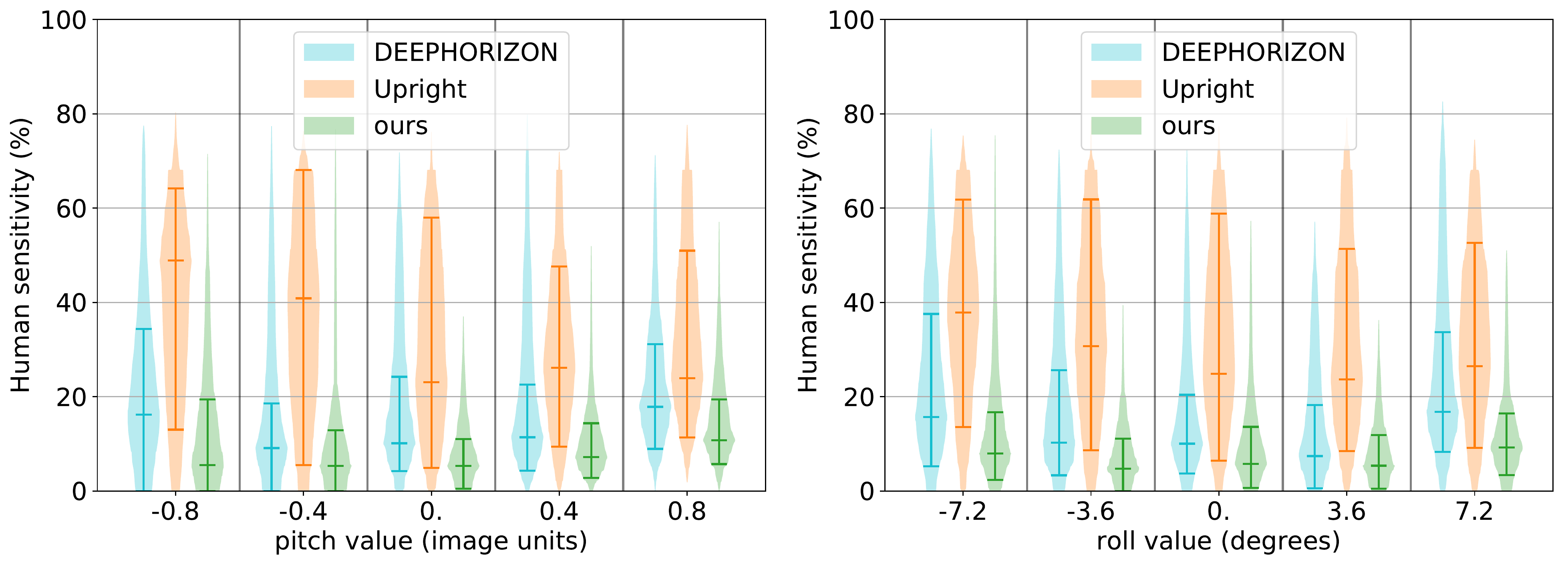}
\begin{tabular}{p{0.5\linewidth}p{0.5\linewidth}}
\hspace{1.5cm}(a) & \hspace{1.3cm}(b)
\end{tabular}
\caption{Performance of our method, Upright and DEEPHORIZON on (a) pitch and (b) roll using our human sensitivity measure. Reported sensitivity takes into account estimation errors on all three parameters at the same time.}
\label{fig:method_human_performance}
\end{figure}

We further evaluate the method we proposed in sec.~\ref{sec:proposed_method} against other learning and non-learning-based methods in terms of the human sensitivity. To obtain a human sensitivity function, we fit a kNN with k=15 to the human sensitivity study results of this section. Specifically, we use error and parameter values on pitch, roll and field of view, yielding a 6 degrees of freedom function. We use the Euclidean distance for neighbor selection and scaled each value according to observed tolerance in the user study: $1{:}0.2$ for pitch in image units, $1{:}12$ for roll in angle and $1{:}15$ for field of view in angle. We convert the reported percentage of human choosing the ground truth values to human sensitivity by mapping the 50-100\% range (confusion-detection) to 0-100\%. Performance on this human sensitivity function on our SUN360 test set for horizon estimation is shown in fig.~\ref{fig:method_human_performance}. Our method has the third quartile of its predictions under 20\% of human sensitivity across the parameter range, systematically lower than every other method. Even though our method was not directly trained using a perceptual loss, we believe this improvement is due to the entropy-based loss being stricter than the perceptual loss (i.e., it penalizes all errors even when they don't affect the realism).

To further assess the perceived accuracy of our method, we ran another user study that showed participants object insertion results using 3 calibration methods simultaneously and asked them to pick the most realistic one. Table~\ref{tab:user_comparison} shows the scores from 2208 votes (32 users $\times$ 69 images).

\begin{table}[!h]
\centering
\vspace{-0.5em}
\begin{tabular}{lll}
\toprule
ours & \cite{Workman2016} & \cite{Lee2014} \\
\midrule
46\% & 31\% & 23\% \\
\bottomrule
\end{tabular}
\vspace{0.5em}
\caption{User study comparing human preference on virtual object insertion. Percentage represents each method's share of votes.}
\label{tab:user_comparison}
\end{table}

%% file: applications.tex
\section{Applications}

We now demonstrate three different uses for camera parameter estimation from a single image: image retrieval, geometrically-consistent object transfer across images, and virtual 3D object insertion.

\paragraph{Image retrieval}

Our technique can be used to retrieve images in large databases based on their camera geometric properties like viewpoint and field of view. To demonstrate this, we estimated the camera parameters using our technique on a subset of 10,000 images randomly selected from the Places2 dataset~\cite{Zhou2017}, computed the intersection of the horizon line with the left and right image boundaries, and ordered images in the dataset based on the L2 distance of these points to ones in the query image. Fig.~\ref{fig:applications_retrieval} presents the 4 closest matches for three query images.

\paragraph{Geometrically-consistent object transfer}

\begin{figure}[!t]
\centering
\includegraphics[width=0.7\linewidth]{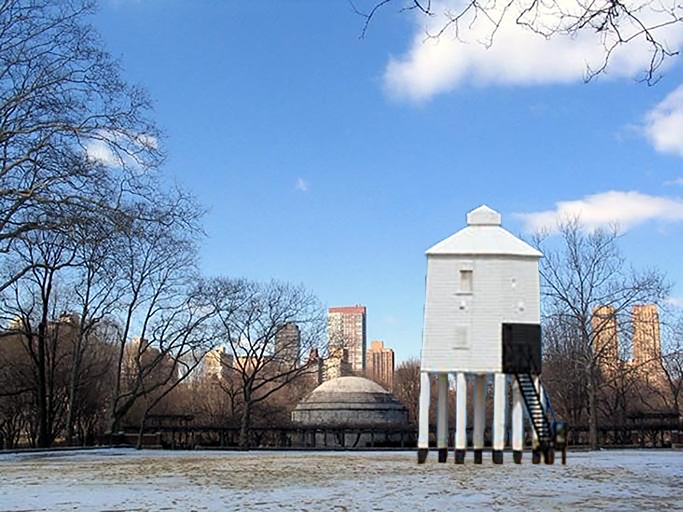}
\caption{The water tower from fig.~\ref{fig:applications_retrieval} pasted onto an image with an automatically detected similar horizon line. Note how the perspective looks right without modification.\vspace{-0.7em}}
\label{fig:applications_2d_compositing}
\vspace{-0.5em}
\end{figure}

Transferring objects from one image to another requires matching the camera parameters~\cite{lalonde-siggraph-07}. While previous techniques required the use of objects of known height in the image in order to infer camera parameters~\cite{lalonde-siggraph-07}, our approach can obtain them from the image itself, and as such can be used to realistically transfer objects from one image to another. One such example is shown in fig.~\ref{fig:applications_2d_compositing}. 

\paragraph{Virtual object insertion}

\newcommand{\voiwidth}{0.43}
\begin{figure}
\centering
\begin{tabular}{@{}cc@{}}
\includegraphics[width=\voiwidth\linewidth]{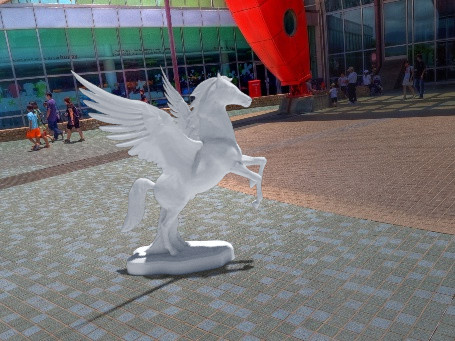} &
\includegraphics[width=\voiwidth\linewidth]{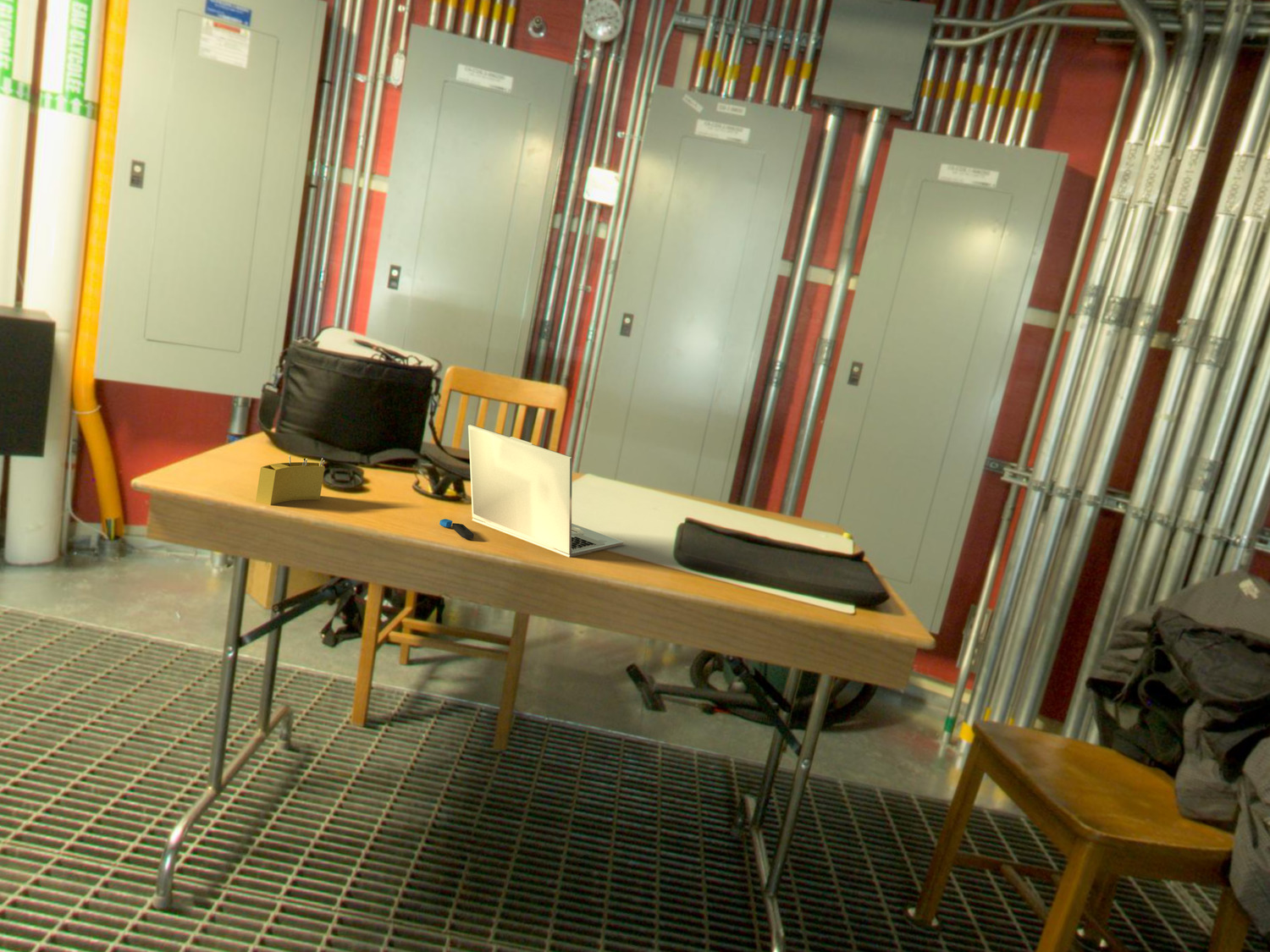} \\
\includegraphics[width=\voiwidth\linewidth]{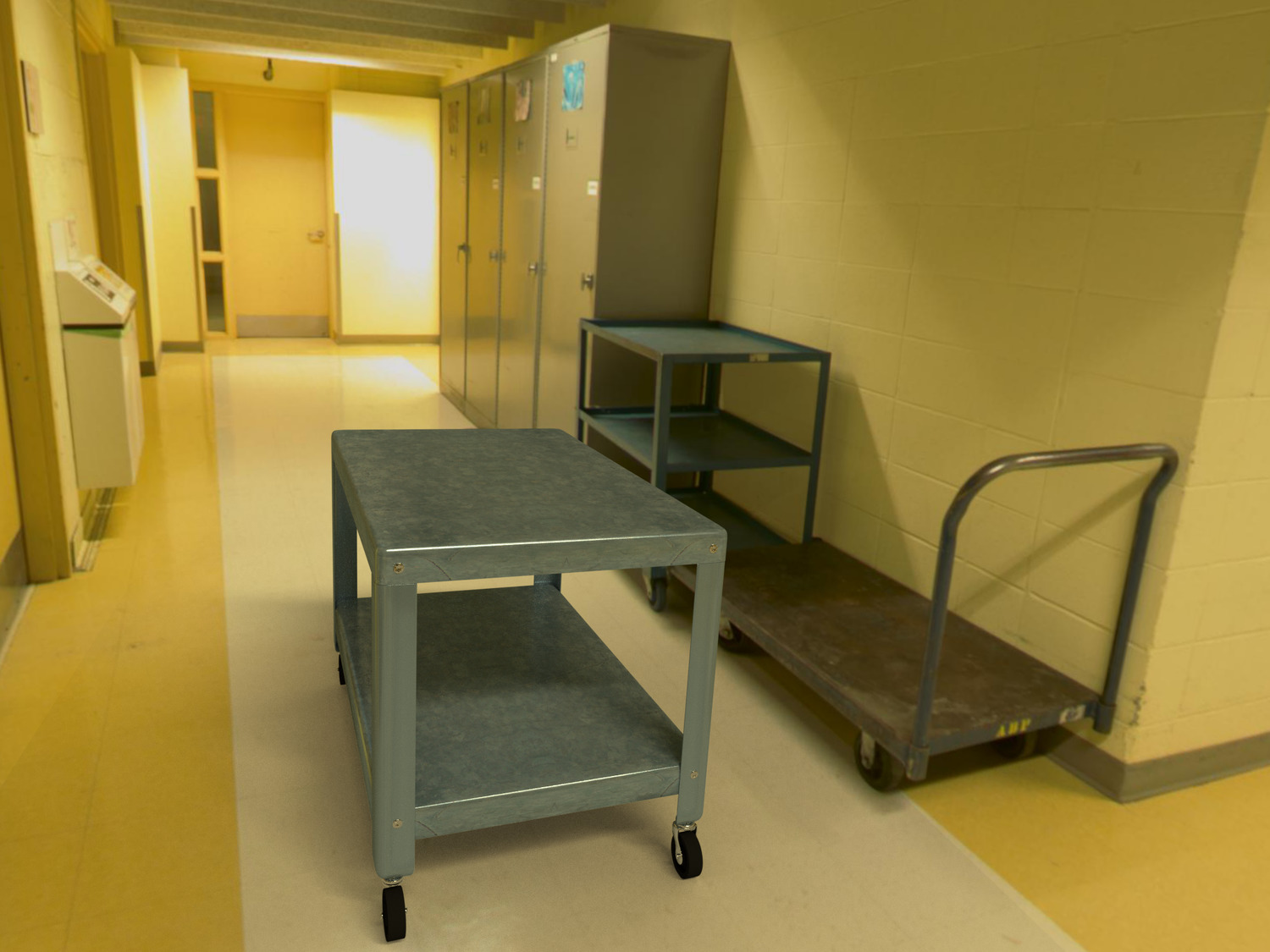} &
\includegraphics[width=\voiwidth\linewidth]{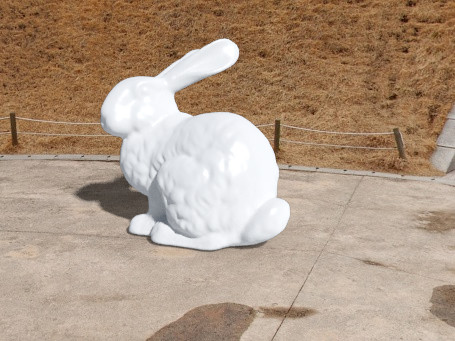}
\end{tabular}
\caption{Examples of virtual object insertions using the camera calibration estimated by our technique. \textbf{More results available in the supplementary material.}\vspace{-0.9em}}
\label{fig:applications_virtual_object_insertion}
\end{figure}

Our approach also enables the realistic insertion of 3D objects in 2D images. As discussed in sec.~\ref{sec:human_sensitivity_analysis}, camera parameters are needed to plausibly align the virtual object with the background image. Given our automatic estimates, the user only needs to select an insertion point in the image and to specify the virtual camera height. Assuming the local scene around the object is a flat plane aligned with the horizon, we can automatically insert a virtual object and demonstrate several such examples in fig.~\ref{fig:applications_virtual_object_insertion}. For these results, the camera height was set to 1.6m and the lighting was automatically estimated by~\cite{Hold-Geoffroy2017,Gardner2017}.

%% file: discussion.tex

\section{Discussion}

In this paper, we present what we believe is the first analysis of human sensitivity on estimation errors for camera pitch, roll and field of view in the context of virtual object insertion. To this end, we performed a large-scale user study on Mechanical Turk, which evaluates how reliably participants were able to distinguish between two images with virtual objects composited with ground truth and distorted camera parameters.
Our study reveals that humans are not always sensitive to large errors, especially when the roll is pronounced, or when the field of view is underestimated. We also present a CNN-based single image calibration estimation method which yields state-of-the-art performance, enabling applications such as image retrieval, geometrically-consistent 2D object transfer, and virtual 3D object insertion. Upon investigation, it was revealed that the learned model is looking for semantically meaningful vanishing lines, making parallels with geometrically-based auto-calibration techniques. Finally, we leverage the user study results to define a distance function based on human perception, which is used to compare our CNN to previous approaches.

Despite this progress, our approach still suffers from a few limitations. While the trained CNN works very robustly in a large number of realistic scenarios, extreme pitch angles (e.g. looking straight down) cannot be represented by the current horizon line parameterization. Furthermore, the perceptual distance function defined in sec.~\ref{sec:cnn-evaluation-perception} could be used as a loss function to train the neural network, putting emphasis on training images where humans are more sensitive to errors. This perceptual loss could also better model the human perception by using photorealistic objects throughout the study. Lastly, the robustness of our model could be coupled with the accuracy of a geometric-based method by taking advantage of the detected semantically meaningful vanishing lines.